\documentclass[letterpaper, 10 pt, conference]{ieeeconf}  

\IEEEoverridecommandlockouts                              

\usepackage{graphics} 
\usepackage{epsfig} 
\usepackage{times} 
\usepackage{amsmath} 
\usepackage{amssymb}  
\usepackage{lipsum}
\usepackage{xcolor}
\usepackage{cite}
\usepackage{hyperref}
\usepackage[utf8]{inputenc}
\usepackage{balance}
\let\labelindent\relax
\usepackage{enumitem}

\usepackage{color}

\newcommand{\st}{\ensuremath{s}}

\newcommand{\ac}{\ensuremath{a}}

\usepackage[most]{tcolorbox}
\usepackage{listings}

\newtcblisting{promptbox}[1]{
  enhanced,
  breakable,
  colback=black!2,
  colframe=black!60,
  title=#1,
  listing only,
  listing options={basicstyle=\ttfamily\small,breaklines=true}
}

\newtcblisting{promptboxOrange}[1]{
  enhanced,
  breakable,
  colback=orange!5,
  colframe=orange!70!black,
  coltitle=black,
  colbacktitle=orange!20,
  title=#1,
  listing only,
  listing options={basicstyle=\ttfamily\small,breaklines=true}
}

\title{\LARGE \bf
GIFT: Generalizing Intent for Flexible Test-Time Rewards 
}

\author{
  Fin Amin$^1$, Nathaniel Dennler$^2$, Andreea Bobu$^2$\\
  $^1$NC State, $^2$MIT CSAIL 
}

\begin{document}

\maketitle
\thispagestyle{empty}
\pagestyle{empty}

\begin{abstract}

Robots learn reward functions from user demonstrations, but these rewards often fail to generalize to new environments. This failure occurs because learned rewards latch onto spurious correlations in training data rather than the underlying human intent that demonstrations represent. Existing methods leverage visual or semantic similarity to improve robustness, yet these surface-level cues often diverge from what humans actually care about. We present Generalizing Intent for Flexible Test-Time rewards (GIFT), a framework that grounds reward generalization in human \textit{intent} rather than surface cues. GIFT leverages language models to infer high-level intent from user demonstrations by contrasting preferred with non-preferred behaviors. At deployment, GIFT maps novel test states to behaviorally equivalent training states via \textit{intent-conditioned similarity}, enabling learned rewards to generalize across distribution shifts without retraining. We evaluate GIFT on tabletop manipulation tasks with new objects and layouts. Across four simulated tasks with over 50 unseen objects, GIFT consistently outperforms visual and semantic similarity baselines in test-time pairwise win rate and state-alignment F1 score.
Real-world experiments on a 7-DoF Franka Panda robot demonstrate that GIFT reliably transfers to physical settings. Further discussion can be found at \href{https://mit-clear-lab.github.io/GIFT/}{https://mit-clear-lab.github.io/GIFT/} 


\end{abstract}

\section{Introduction}\label{sec:intro}

\begin{figure}[t]
  \centering 
  \includegraphics[width=\columnwidth]{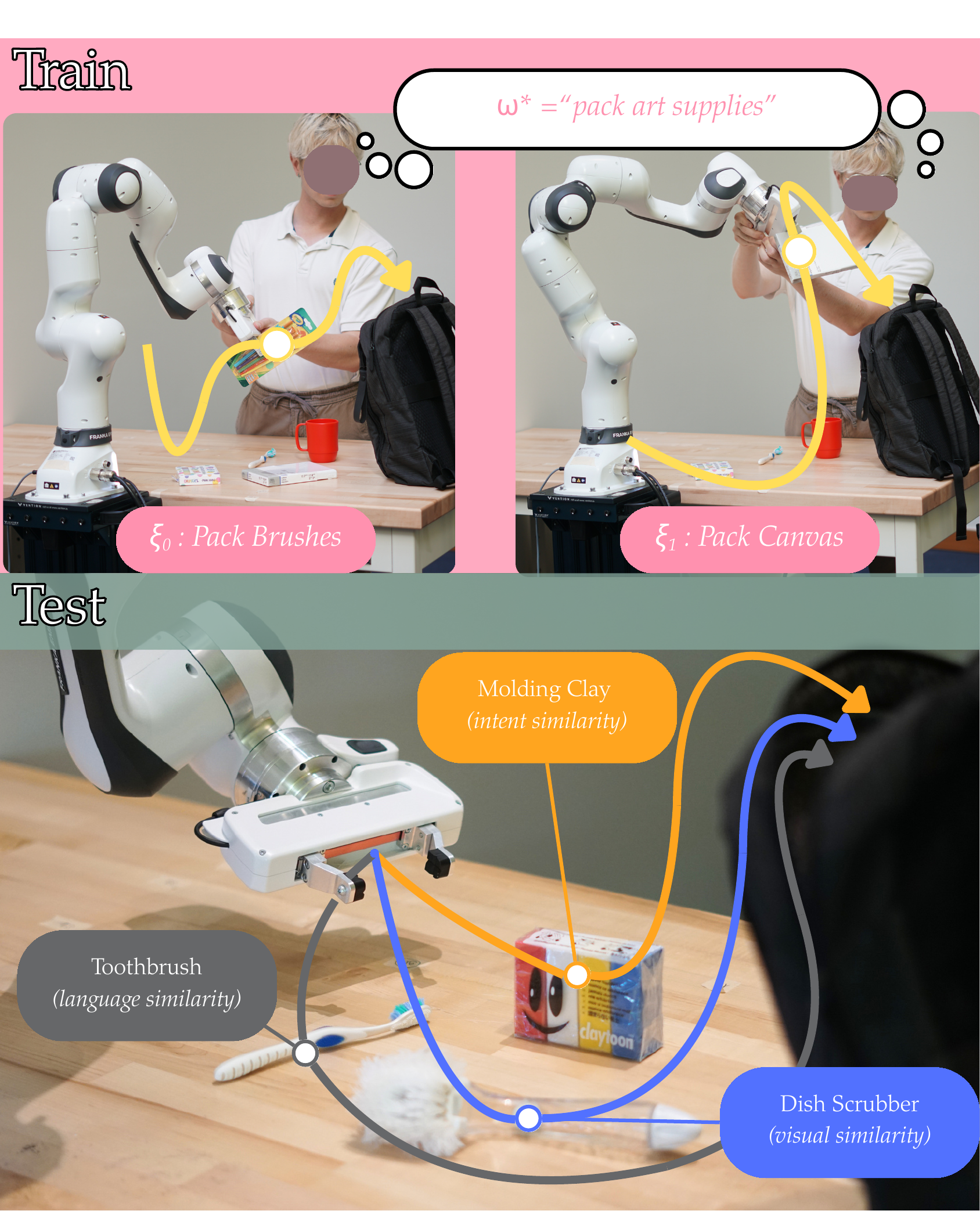}
  \vspace{-8mm}
  \caption{\textbf{Top.} 
  During training, the robot gets task demonstrations (loading a paintbrush) and uses them to infer the user's \textit{high-level intent} (``load art supplies'').
  \textbf{Bottom.} At test time, the robot encounters three unseen objects. GIFT uses the inferred intent to correctly identify that the molding clay is relevant. In contrast, visual-similarity baselines incorrectly prefer the dish scrubber due to its appearance, and language-similarity baselines make an analogous mistake (``\textit{tooth\textbf{brush}}'' and ``\textit{paint\textbf{brush.}}'')}
  \label{fig:behavior}
  \vspace{-7mm}
\end{figure}

Imagine a human teaching a robot how to pack their bag for an art class (Fig.~\ref{fig:behavior}). The human might demonstrate placing a paintbrush and sketchbook in the bag, expecting the robot to understand their goal of gathering art supplies. Later, when preparing for a different art session, the human would naturally trust the robot to assist them autonomously with packing the right supplies---whether it's the same paintbrush and sketchbook, or instead molding clay, an easel, or any other materials needed for the task.

Unfortunately, while robots can learn reward functions from user demonstrations~\cite{finn2016guided,fu2018AIRL}, these reward functions rarely generalize to new scenarios. Instead, learned rewards often latch onto spurious correlations in the training data~\cite{agrawal2022task} (e.g., reward functions learn to ``pack paintbrush'') rather than capturing the underlying high-level intent (e.g., a reward function to ``pack art supplies''). As a result, when robots encounter new scenarios at test time, the learned reward may fail catastrophically, causing the robot to either ignore important task aspects or fixate on irrelevant ones. 
Prior work has attempted to improve robustness through data augmentation based on visual~\cite{DVDLearningGeneralizableRewards} or language~\cite{semantically_controllabe_augmentations} similarity, but these measures often diverge from what humans actually care about when defining task success~\cite{bobu2023SIRL}.

Our key insight is that \textit{state similarity is driven by people's high-level task intent, not by surface-level visual or language cues}. In our example in Fig.~\ref{fig:behavior}, a vision-based model might judge a scrubber to be most similar to a paintbrush because they look alike, while a language model might instead group a paintbrush with a toothbrush due to lexical similarity. Neither captures what actually matters for the task: under the intent ``pack art supplies,'' a paintbrush is closer to molding clay. In addition, different intents induce different notions of similarity: if the intent were instead ``set aside painting supplies,'' then paintbrush, watercolor, and easel would be similar to the intention, while molding clay and sketchbook would be dissimilar. Without conditioning on intent, the robot cannot know which notion of similarity is appropriate, and thus risks generalizing along the wrong notion of similarity.

\begin{figure*}[ht]
    \centering 
   \includegraphics[width=1.45\columnwidth, trim={0 175 0 0}, clip]{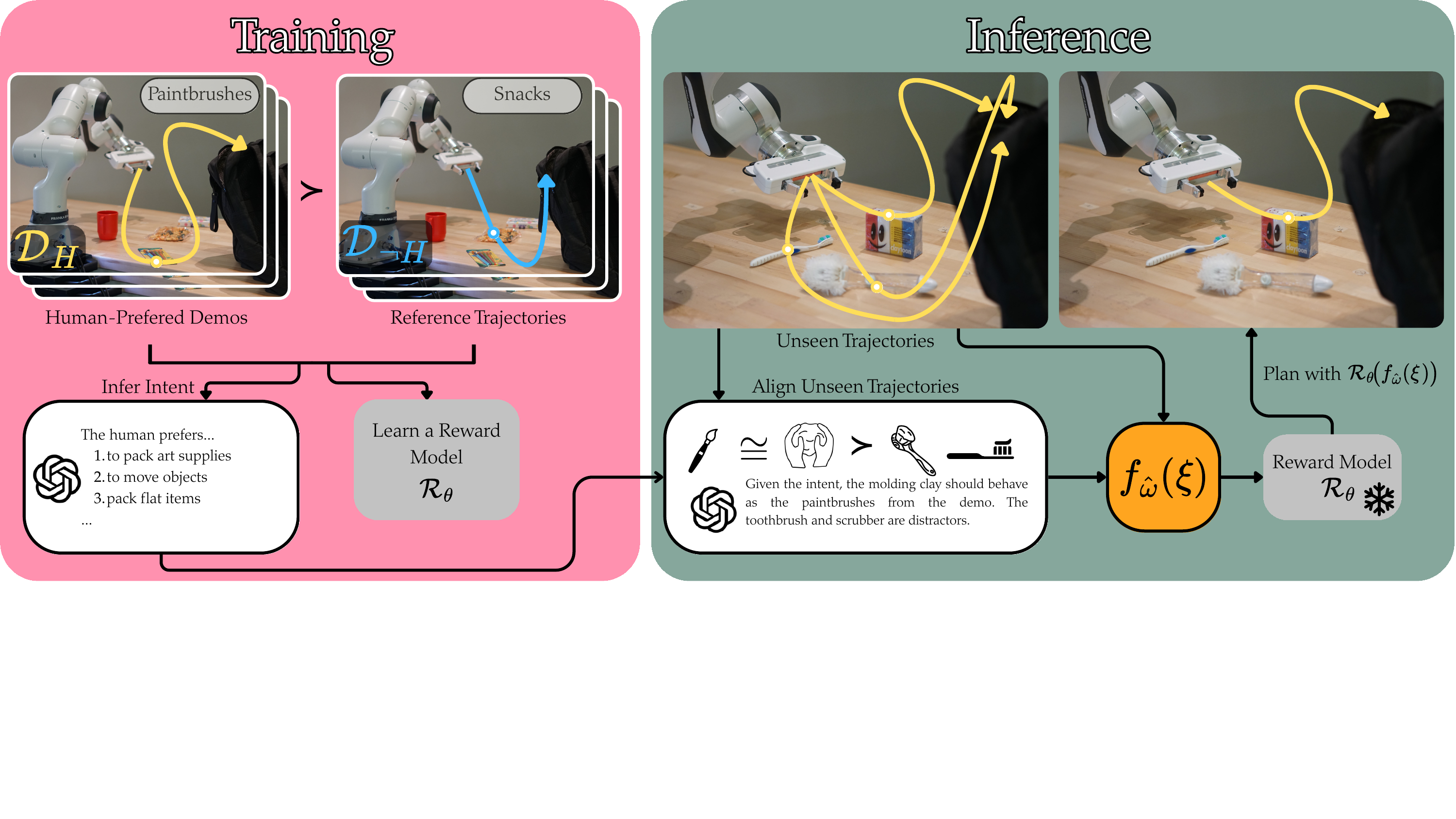} 
   \vspace{-8mm}
    \caption{\textbf{G}eneralizing \textbf{I}ntent for \textbf{F}lexible \textbf{T}est-Time rewards. \textbf{Left.} GIFT infers the human's intent given pairs of human-preferred demonstrations and reference trajectories. \textbf{Right.} During inference, GIFT deduces which objects in the unseen states should behave as objects in the training states. Afterwards, the unseen state components are aligned to training states so that the reward function learned before deployment can be used for planning.}
    \label{fig:GIFT_Arch}
    \vspace{-5mm}
\end{figure*}

To address this issue, we introduce \textbf{G}eneralizing \textbf{I}ntent for \textbf{F}lexible \textbf{T}est-Time rewards (GIFT), a framework that grounds reward generalization in high-level intent. We leverage language models' (LMs) commonsense reasoning abilities in two stages. First, we infer the human's high-level intent by prompting LMs with contrasting pairs of user-preferred trajectories and non-preferred trajectories. Second, at test time, the inferred intent conditions another LM call that aligns novel test states to semantically similar training states. This lets the robot reuse its learned reward function without retraining. In experiments with a 7-DoF Franka Panda robot, both in simulation and the real world, we show that GIFT substantially outperforms low-level visual and language baselines on test-time reward generalization.

\section{Related Work}\label{sec:related_work}

\smallskip\noindent\textbf{Learning Rewards from Humans.}
Robots commonly learn human goals from demonstrations~\cite{abbeel2004apprentice,finn2016guided,fu2018AIRL}, preferences~\cite{christiano2017prefs,myers2022learning}, or corrections~\cite{mehta2024unified,bobu2018learning,bajcsy2018learning}. Language has recently emerged as a powerful medium for reward learning, enabling free-form text or natural language critiques to refine reward inference~\cite{sumers2021lingrewards,ElicitingHumanPref}. Other methods leverage LMs as priors to propose interpretable reward features or to ground abstract task objectives directly into reward functions~\cite{huang2022zeroshot,ma2023eureka,li2023lampp,peng2024algae}. While these approaches make it easier for non-experts to convey their intent without hand-engineering reward functions, the resulting models often overfit to surface correlations in limited data, leading to \textit{reward misspecification}---where the learned reward captures proxies of success rather than the user’s true objective~\cite{amodei2016concrete,bobu2024ARHR}. GIFT aims to avoid reward misspecification by interpreting demonstrations as expressions of high-level \textit{intent} that can be inferred and transferred across domains to enable flexible test-time reward generalization.

\smallskip\noindent\textbf{Test-Time Adaptation in Robots.}
Reward learning approaches often require users to teach the robot again when it encounters a new environment. Collecting data for every new environment can be an onerous process for users~\cite{FERL,LearningStructure,CLEA,rosid}, so other techniques aim to transfer the robot's understanding of user preferences to unseen contexts at test time~\cite{peng2023dfa,forseysmerek2025contextmatterslearninggeneralizable,peng2024plga,LearningToIdentifyNewObjects,DVDLearningGeneralizableRewards,TargetProjection}. Test-time adaptation is often achieved by using large pretrained models to directly map from test-time observations to data seen during training by using textual correspondence between objects~\cite{LearningToIdentifyNewObjects, ROBO_ABC}, visual correspondence between videos~\cite{DVDLearningGeneralizableRewards}, or by projecting test images to the training manifold of images~\cite{TargetProjection}.  In contrast to previous approaches that perform mappings that directly reflect the data distribution, GIFT aims to leverage higher-order representations of \textit{user intention} that are robust to distribution shifts.



\section{Problem Formulation}

We consider the problem of adapting a robot's behavior to align with human objectives under distribution shift. 
Our goal is to enable a learned reward function to generalize across distribution shifts by grounding it in the human’s high-level intent, rather than correlations present in the training domain.

\smallskip
\noindent\textbf{Markov Decision Processes.} We model our setting as a Markov Decision Process (MDP)~\cite{puterman2014markov} $\mathcal{M} = \langle \mathcal{S}, \mathcal{A}, \mathcal{T}, \mathcal{R} \rangle$, where $\mathcal{S}$ is the state space (e.g., robot joint poses, object poses), $\mathcal{A}$ the action space, $\mathcal{T}: \mathcal{S} \times \mathcal{A} \times \mathcal{S} \rightarrow [0,1]$ the transition dynamics, and $\mathcal{R}: \mathcal{S} \times \mathcal{A} \rightarrow \mathbb{R}$ the reward function. We denote trajectories as state-action sequences $\xi=(\st_0, \ac_0, \dots, \st_{T-1},\ac_{T-1}, \st_T)$. The robot aims to find a policy $\pi: \mathcal{S} \rightarrow \mathcal{A}$ such that the trajectories induced by $\pi^*$ maximize cumulative rewards, $\pi^* = \arg\max_{\pi} \mathbb{E}_{\xi \sim \pi}[\mathcal{R}(\xi)]$.

\smallskip
\noindent\textbf{Reward Learning from Human Input.}
The reward is not known a priori, as it reflects the human's internal preference over how the robot should behave. We model this preference as a parameterized reward function $\mathcal{R}_\theta(\xi) = \sum_{\st_t \in \xi} \mathcal{R}_\theta(\st_t)$, where $\theta$ are learnable parameters. Because hand-specifying $\mathcal{R}_\theta$ is difficult, the robot can instead \emph{learn} it from human input such as demonstrations, preferences, or corrections. 

The robot's goal is to find parameters $\theta$ such that the reward $\mathcal{R}_\theta$ captures the human's underlying intent. This can be achieved through various approaches including inverse reinforcement learning (IRL)~\cite{ziebart2008maximum,finn2016guided}, preference learning~\cite{christiano2017prefs}, learning from corrections~\cite{bajcsy2018learning}, etc. Our method (Sec.~\ref{sec:method}) is agnostic to the reward learning algorithm. For concreteness, we focus our exposition on the IRL setting, where the algorithm seeks $\theta$ such that human-demonstrated trajectories ${\mathcal{D}_{H}} = \{\xi_i\}_{i=1}^N$ are more likely under $\mathcal{R}_\theta$ than alternative trajectories  $\xi \in \Xi$, where $\Xi$ denotes the set of all possible trajectories. In this formulation, demonstrations serve as evidence of desired behavior, and reward learning amounts to recovering the preference $\mathcal{R}_\theta$ that best explains them.


\smallskip
\noindent\textbf{Distribution Shift.}
Standard reward learning assumes that training and test trajectories are drawn from the same distribution. In practice, however, robots are deployed in environments where test-time states differ from those observed in training---due to new objects, layouts, or contexts. This distribution shift leads to two major failure modes: \textit{reward misspecification} and \textit{out-of-distribution misgeneralization}.

First, when demonstrations $\mathcal{D}^{\mathrm{train}}$ cover only a subset of the state space $\mathcal{S}^{\mathrm{train}} \subseteq \mathcal{S}$, the robot may recover a reward function $\mathcal{R}_\theta$ that explains the specific training data but fails to capture the human’s true \textit{high-level} intent. For example, demonstrations of keeping a coffee cup away from a laptop may be misinterpreted as ``avoid laptops'' rather than the intended ``avoid liquids near water-sensitive objects.'' In such cases, the learned reward $\mathcal{R}_{\theta}$ encodes spurious correlations tied to $\mathcal{S}^{\mathrm{train}}$, diverging from the intended reward $\mathcal{R}_{\theta^*}$. This reflects a \emph{reward misspecification} problem where partial observability and limited coverage of the demonstrator's intent lead to learning proxy objectives rather than the true underlying preference~\cite{amodei2016concrete,bobu2024ARHR}.

Second, even when the intended preference is correctly recovered within the training domain, applying $\mathcal{R}_\theta$ to novel test states $\mathcal{S}^{\mathrm{test}}$ can produce unreliable behavior. The learned reward function, having been optimized to explain demonstrations involving specific objects and configurations, may respond unpredictably to states containing unseen elements. For instance, a reward function trained on demonstrations involving a \emph{paintbrush} may assign arbitrary values to states containing \emph{molding clay}, even when both objects serve the same functional role in the task. This represents an \emph{out-of-distribution misgeneralization} failure where the learned reward produces unreliable outputs on novel inputs, regardless of the underlying parameterization~\cite{shimodaira2000improving,agrawal2022task}.

These two challenges illustrate why standard reward learning fails to generalize under distribution shift: directly applying $\mathcal{R}_\theta$ to $\mathcal{S}^{\mathrm{test}}$ may assign arbitrary or misleading values to novel states not covered by $\mathcal{S}^{\mathrm{train}}$. Our goal is to enable reward functions to adapt to such shifts without retraining by grounding them in the human’s high-level intent.

\section{Generalizing Intent for Flexible Test-Time Rewards (GIFT)}\label{sec:method}

GIFT achieves test-time reward generalization by defining a \textit{state similarity function conditioned on the human's high-level intent}. This function allows us to map novel test states into the training domain and reuse the reward function without retraining. Fig.~\ref{fig:GIFT_Arch} presents an overview of our framework.

\subsection{The GIFT Framework}

\vspace{-1mm}
\noindent\textbf{Inferring High-Level Intent.}
To align states with an intent-conditioned similarity function, we first infer the human's underlying intent. 
We leverage LMs' reasoning capabilities and provide it with:
(i) human-preferred demonstrations $\mathcal{D}_{H}$ and (ii) reference trajectories $\mathcal{D}_{\neg H} := \{ \xi_i \sim \Xi \}_{i=1}^N$ (e.g., rollouts from a nominal controller or other non-preferred behavior) in the same scenes. By contrasting these behaviors~\cite{peng2024algae}, the LM outputs a natural language summary of the high-level intent that explains the difference between $\mathcal{D}_{H}$ and $\mathcal{D}_{\neg H}$ (Fig.~\ref{fig:GIFT_Arch} left). 
We denote this process as
\begin{equation*}
\omega^* \approx \hat{\omega} \;\triangleq\; J\!\big(\mathcal{D}_{\mathrm{H}},\, \mathcal{D}_{\neg H}\big),
\end{equation*}
where $\omega^*$ is the human's true high-level intent, $\hat{\omega}$ is our estimate, and $J$ is the intent estimator instantiated as an LM.

\smallskip
\noindent\textbf{Intent-Conditioned State Similarity.}
Having recovered the intent $\hat{\omega}$, we now describe how it guides our notion of state similarity. Our key insight is that states should be treated as similar based on the human's high-level intent rather than low-level visual or language features. We formalize this with an intent-conditioned kernel:
\begin{equation*}
\mathcal{K}(s,s' \mid \omega):\; \mathcal{S} \times \mathcal{S} \to [0,1],
\end{equation*}
where higher values denote greater similarity and $\omega$ determines what ``similar'' means for the task. Existing methods in the literature typically define similarity using unimodal representations such as low-level visual~\cite{DINO} or language~\cite{BERT} features. In contrast, we posit that high-level intent $\omega$ inherently captures cross-modal semantic relationships---functional goals that transcend specific visual or linguistic instantiations---that better reflect task-relevant similarity.

For example, consider a training scene where a \emph{paintbrush} is packed into a backpack and a test scene with different objects (Fig.~\ref{fig:behavior}). A vision-based kernel might show high similarity between a \emph{scrubber} and a \textit{paintbrush} because they are visually alike, while a language-based kernel might instead show a high similarity between a \textit{toothbrush} and a  \emph{paintbrush} due to lexical similarity. 
In contrast, our intent-conditioned kernel recognizes that \emph{molding clay} and \textit{paintbrush} are similar when conditioned on the intent ``pack art supplies,'' preserving the demonstrator’s high-level intent.

Different high-level intents induce different measures of similarity over the same set of states. For example, under the intent ``pack art supplies,'' states containing a \textit{paintbrush}, \textit{molding clay}, or \textit{sketchbook} are treated as equivalent, while under ``set aside painting supplies,'' the relevant cluster shifts to \textit{paintbrush}, \textit{watercolor}, and \textit{easel}, with \textit{molding clay} and \textit{sketchbook} now irrelevant. Without conditioning on intent, a robot cannot distinguish which notion of similarity is appropriate and may generalize incorrectly.

To operationalize the kernel $\mathcal{K}$, we use LMs as implicit similarity functions: given the estimated intent $\hat{\omega}$ and descriptions of two states, the LM leverages common-sense priors to decide how similar the states are (Fig.~\ref{fig:GIFT_Arch} right). This enables us to align novel test states to behaviorally equivalent training states in a way that preserves the underlying human intent, rather than spurious correlations. While our implementation uses binary relevance decisions for simplicity, the framework could naturally extend to continuous similarity by querying the LM to rate relevance on a continuous scale. 

\smallskip
\noindent\textbf{Aligning Test States.}
Once we infer the human’s intent $\hat{\omega}$, we use it to align test states to their nearest intent-equivalent training states. Let $\mathcal{S}^{\mathrm{train}}$ be a dataset of states collected from training scenes. For a novel state $s'\in \mathcal{S}^{\mathrm{test}} $, we define an intent-conditioned alignment operator $f_{\hat\omega}$: 
\begin{equation*}
f_{\hat\omega}(s') \;\triangleq\; \operatorname*{argmax}_{s\in \mathcal{S}^{\mathrm{train}}}\; \mathcal{K}\!\big(s,\,s'\mid \hat\omega\big),
\end{equation*}
which maps\footnote{For simplicity, we write this as operating over full states, though in practice the alignment applies to the semantic components of the state (e.g., object identity) while preserving continuous dimensions such as robot pose.} $s'$ to the nearest intent-equivalent training state identified by the kernel $\mathcal{K}$.
We extend this operator to trajectories $\xi' = (s'_0, \ldots, s'_T)$ by applying it state-wise, aligning each state independently:
\begin{equation*}
f_{\hat\omega}(\xi') \;\triangleq\; \big(f_{\hat\omega}(s'_0), \ldots, f_{\hat\omega}(s'_T)\big).
\end{equation*}
This produces an aligned trajectory $f_{\hat\omega}(\xi')$ whose semantic components correspond to training states in $\mathcal{S}_{\mathrm{train}}$. Intuitively, the alignment operator projects each novel test state into the training domain along intent-relevant dimensions, ensuring that subsequent reward evaluation depends on behaviorally equivalent states.

In our running example, suppose a test scenario involves packing molding clay. Under the inferred high-level intent $\hat\omega=$ ``pack art supplies,'' the alignment operator $f_{\hat{\omega}}$ maps molding clay to its training equivalent paintbrush, yielding an aligned trajectory in the training domain whose reward evaluation reflects the intended goal.

\smallskip
\noindent\textbf{Test-Time Reward Generalization.}
Let $\mathcal{R}_\theta$ be the reward learned once on training data. At test time, we evaluate unseen trajectories by first aligning them into the training domain and then applying $\mathcal{R}_\theta$:
\begin{equation*}
\widetilde{\mathcal{R}}_\theta(\xi' \mid \hat{\omega}) \;\triangleq\; \mathcal{R}_\theta\!\big(f_{\hat{\omega}}(\xi')\big).
\end{equation*}
In this formulation, the alignment operator $f_{\hat{\omega}}$ carries the ``burden'' of test-time generalization, while the learned reward $\mathcal{R}_\theta$ itself remains unchanged. At deployment, planning, or policy selection is performed using the fixed reward on aligned trajectories:
\begin{equation*}
\pi^{\star} = \arg\max_{\pi}\; \mathbb{E}_{\xi' \sim \pi}\!\left[\,\widetilde{\mathcal{R}}_\theta(\xi' \mid \hat{\omega})\,\right].
\end{equation*}
Because the alignment is intent-conditioned rather than purely visual- or language-conditioned, optimization proceeds over states that are similar in \emph{meaningful, preference-driven} ways, not merely along surface cues.


\smallskip

\subsection{GIFT Implementation Details}\label{sec:gift_implementation}
\smallskip
\noindent\textbf{Parameterization of the Intent Estimator.}
We estimate the human's intent, $\hat{\omega}_{}$, via a contrastive LM call,
\begin{equation*}
\hat{\omega}_{}=J_{\mathrm{LM}}(\mathcal{D}_{\mathrm{H}}, \mathcal{D}_{\mathrm{\neg H}}),
\end{equation*}
where $\mathcal{D}_{\mathrm{H}}$ are human-provided demonstrations and $\mathcal{D}_{\mathrm{\neg H}}$ are reference trajectories from a nominal controller (e.g., shortest-path planner) in the same scenes. 
To enable language-based reasoning, we follow Peng et al.~\cite{peng2024algae} and represent trajectories in terms of human-interpretable feature values (e.g., end-effector position, distance to objects, gripper orientation) together with natural-language descriptions of what each feature measures. The LM is thus given both (i) structured numeric feature traces and (ii) semantic descriptions of those features, allowing it to reason abstractly about behavior rather than raw sensor values. We prompt the LM to identify the higher-level intent that distinguishes $\mathcal{D}_{\mathrm{H}}$ from $\mathcal{D}_{\mathrm{\neg H}}$, producing abstract summaries such as ``avoid liquids near water-sensitive electronics'' rather than instance-specific descriptions like ``avoid coffee above laptop.''


\smallskip
\noindent\textbf{Intent-Guided Alignment.}
For each test state $s'$, we use the inferred intent $\hat{\omega}$ to determine which training state (if any) is behaviorally equivalent. We implement the alignment operator $f_{\hat{\omega}}$ via an LM call that receives $\hat{\omega}$ as context and is prompted to map the semantic components of $s'$ to their intent-equivalent elements in the training set.

The LM performs a binary relevance judgment for each semantic element of the state (e.g., object identity). If an element is relevant under the intent $\hat{\omega}$, it is mapped to the corresponding training element that fulfills the same intent-relevant role ($\mathcal{K}(s, s' \mid \hat{\omega}) = 1$); otherwise, it is mapped to a \textit{distractor} class ($\mathcal{K}(s, s' \mid \hat{\omega}) = 0$). Formally, this produces a mapping $f_{\hat{\omega}}: \mathcal{S}^{\mathrm{test}} \to \mathcal{S}^{\mathrm{train}} \cup \{\textit{distractor}\}$ conditioned on $\hat{\omega}$ and the scene context, which aligns only the semantic components while preserving continuous ones such as robot pose. Conceptually, this behaves as a \emph{target projection}~\cite{TargetProjection}: at test time, we make new states look like appropriate training states. We use GPT-4o~\cite{GPT4} for all LM calls. A sensible question is whether the alignment could distort reward evaluation on new test states. We investigate this via our alignment error bound located in the Appendix (see Sec. \ref{sec:error_bound}).



\smallskip
\noindent\textbf{Implementation of $\mathcal{R}_\theta$.} GIFT is agnostic to the reward parameterization and learning algorithm. 
In our experiments, we adopt Maximum Entropy IRL~\cite{ziebart2008maximum}, which models trajectory probabilities as $p_\theta(\xi) \propto \exp(\mathcal{R}_\theta(\xi))$ and seeks parameters $\theta$ that maximize the likelihood of given demonstrations. Following Finn et al.~\cite{finn2016guided}, we learn $\mathcal{R}_\theta$ from a set of training demonstrations collected in multiple object configurations.





We parameterize the reward as a linear function over trajectory features: $\mathcal{R}_\theta(\xi) = \theta \cdot \phi(\xi)$, where $\phi: \Xi \to \mathbb{R}^d$ extracts time-aggregated features (e.g. end-effector distance to objects). This linear parameterization enables efficient learning, though GIFT's framework supports more expressive function approximators such as neural networks. Crucially, at deployment the learned reward $\mathcal{R}_\theta$ and feature extractor $\phi$ remain fixed. All test-time generalization to novel objects and scenes occurs through the intent-conditioned alignment operator $f_{\hat{\omega}}$ (Sec.~\ref{sec:method}), with no retraining required.


\begin{figure}[t]
    \centering
    \includegraphics[width=0.775\linewidth]{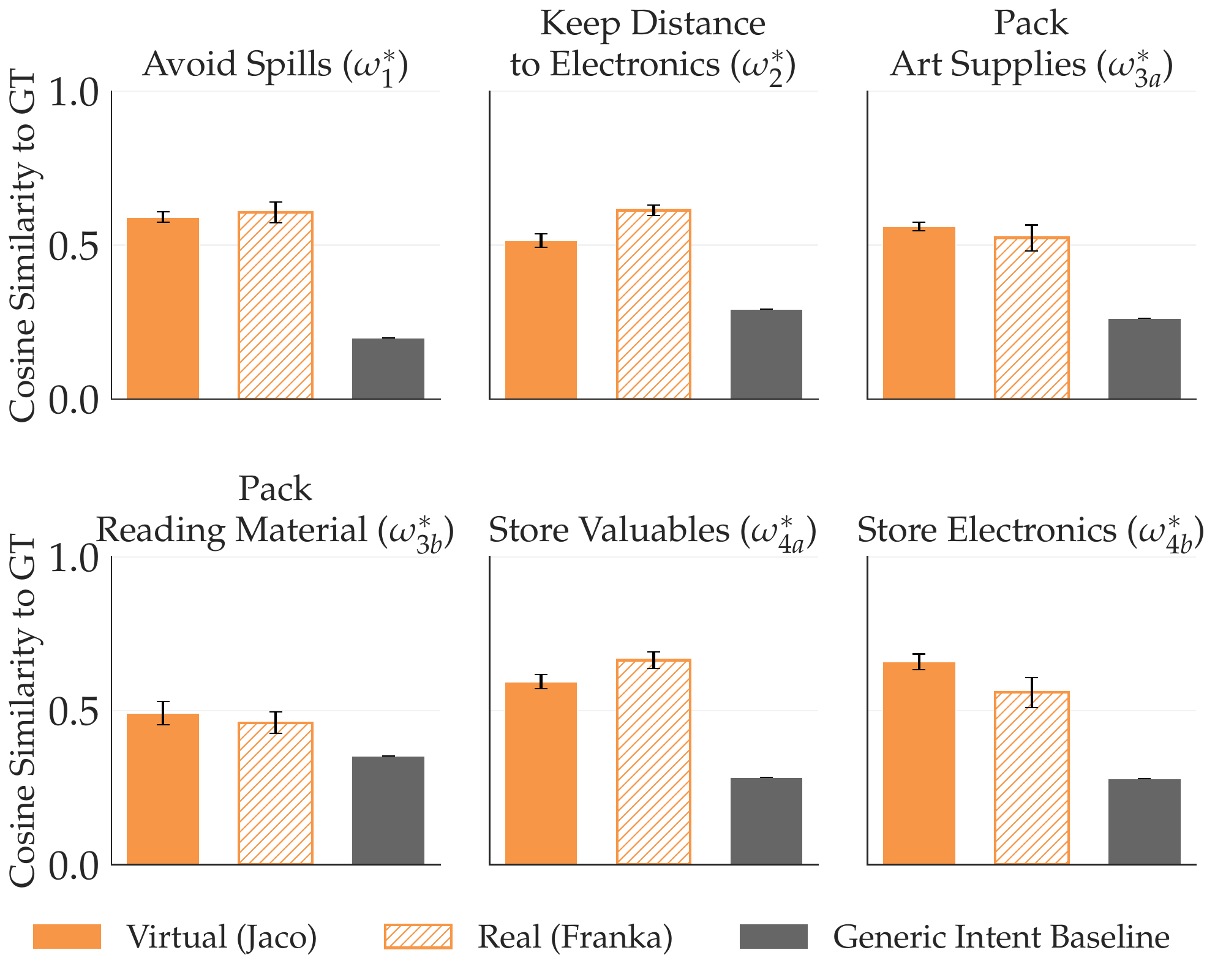}
    \vspace{-4mm}
    \caption{\textbf{Similarity Between LM-Inferred and Ground Truth Intent.} We gave the LM 3 demonstration pairs from a virtual Jaco robot and a real-world Franka robot, and tasked it with deducing the human's intent. We found that LMs produced an acceptable conditioning variable for alignment.}
    \label{fig:lm1_intent}
    \vspace{-5mm}
\end{figure}

\section{Experiments}\label{sec:experiments}

GIFT is based on the idea that high-level intent enables better test-time generalization compared to low-level vision or language features. We explore the following research question to quantify the benefits of GIFT:

\begin{enumerate}[label=\textbf{RQ\arabic*.}, leftmargin=*, align=left]
    \item Does high-level \textit{intent similarity} lead to better reward performance at test-time compared to low-level \textit{visual} or \textit{language} similarity?
    \item When do low-level \textit{visual} or \textit{language} features fail to generalize at test-time?
    \item Does GIFT transfer to robots in the physical world?
\end{enumerate}

We conduct experiments to evaluate these research questions using 7-DoF robot arms across four tabletop manipulation tasks in simulation and the real world. We also document the prompts we used for both GIFT and LM$_{\mathrm{No\; Intent}}$ in the Appendix (see Sec.\ref{sec:prompt_templates}).

\smallskip\noindent\textbf{Tasks.} Our simulated experiments used a 7DoF Jaco robot arm in the PyBullet simulator~\cite{coumans2019}, and our real-world experiments used the 7-DoF Franka Research Robot. Each task corresponds to a distinct ground-truth high-level intent. The tasks in order of increasing complexity are:  
    \textbf{(1) Place Mug.} ${\omega^*_{1}}$: \textit{avoid carrying fluids near water-sensitive objects.}
    \textbf{(2) Sweep Spill.} ${\omega^*_{2}}$: \textit{sweep paper-based items away from the spill}
    \textbf{(3) Pack Backpack$^\star$.} ${\omega^*_{3a}}$: \textit{pack art supplies}; ${\omega^*_{3b}}$: \textit{pack reading material}
    \textbf{(4) Store Into Drawer$^\star$.} ${\omega^*_{4a}}$: \textit{store valuables}; ${\omega^*_{4b}}$: \textit{store electronics}.
Tasks 3 and 4 have multiple possible intents to showcase GIFT's ability to recover diverse intent-conditioned similarity functions.

Across our tasks, we used a dataset of over 50 objects in total, which we split into a distinct training set and test set for each task. The test environments contain different objects from the training environments.

\begin{figure}[t]
    \centering
    \includegraphics[width=0.75\linewidth]{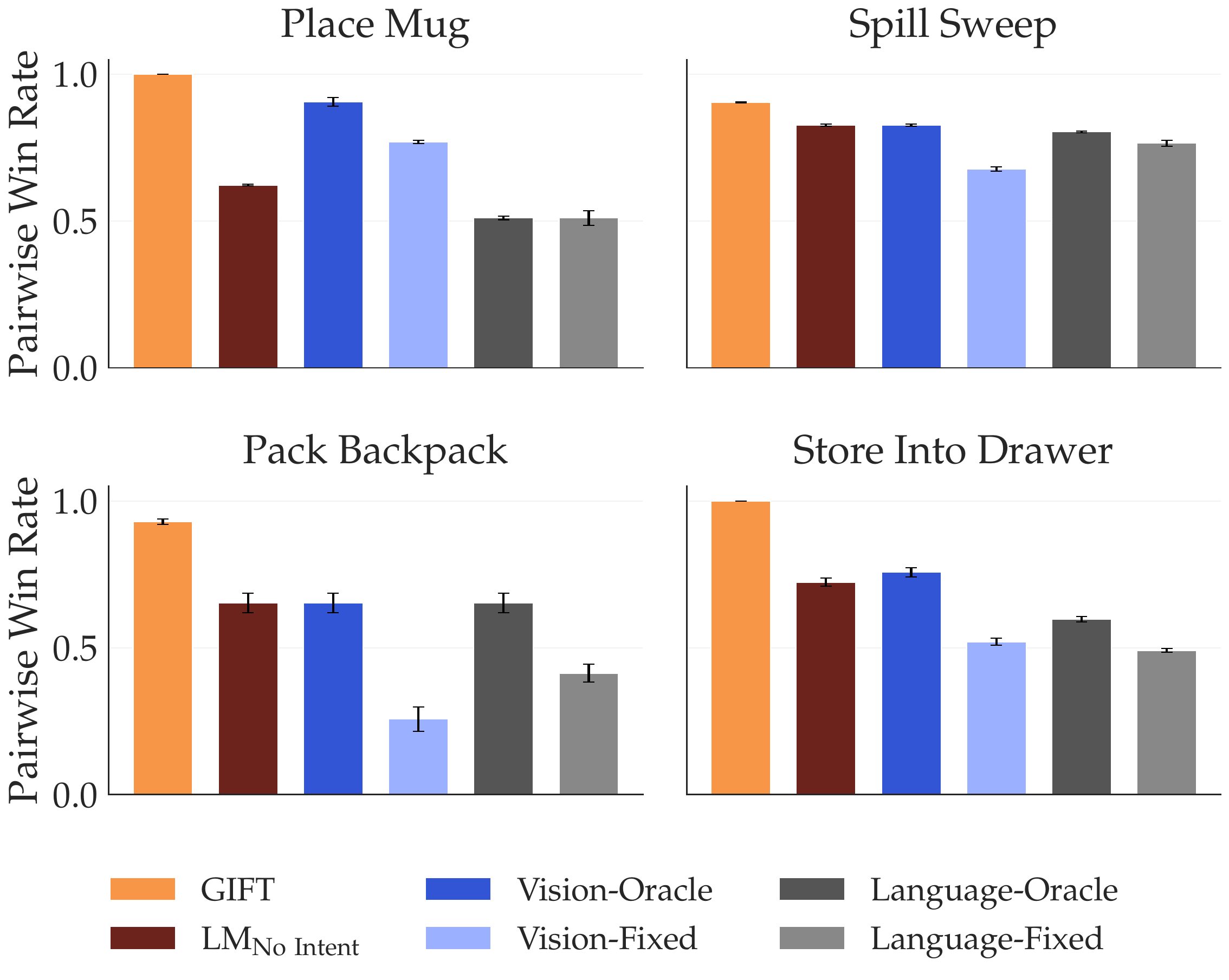}
    \vspace{-3mm}
    \caption{\textbf{Test-Time Pairwise Win Rate.} Across our tasks, the rewards learned via GIFT achieved a higher win rate than all other baselines. These results were aggregated over 250 trajectory pairs per scene, with randomization of various state components along with random sampling from a pool of over 50 unseen objects. Black bars denote standard errors. }
    \label{fig:winrate}
    \vspace{-2mm}
\end{figure}

\noindent\textbf{Sanity Check.} Before evaluating the effectiveness of GIFT, we first verify that GIFT's LM-inferred intents are reasonable estimates of ground-truth intents. We evaluate intent accuracy for each task using the cosine similarity between ground truth intent, ${\omega^*}$, and the intent estimated by $J_{\mathrm{LM}}$. We perform the intent estimation process ten times for each of the four tasks to calculate the mean and standard error. A baseline intent of predefined, task-relevant goals (e.g., “the human prefers to move objects”) serves as a control. To demonstrate flexibility across human preferences, we also evaluated across the two ground-truth intents for \textbf{Pack Backpack} ($\omega^*_{3a}$ and $\omega^*_{3b}$) and \textbf{Store Into Drawer} ($\omega^*_{4a}$ and $\omega^*_{4b}$).

Fig.~\ref{fig:lm1_intent} shows cosine similarity between LM-inferred intent and the ground-truth intent for both simulated (Jaco) and physical (Franka) demonstrations. We found that across all tasks, LM-inferred intents showed higher cosine similarity with the ground truth intent than the generic intent baseline, indicating that the intent-prediction step of GIFT generates reasonable intents.

\begin{figure}[t]
    \centering
    \includegraphics[width=0.75\linewidth]{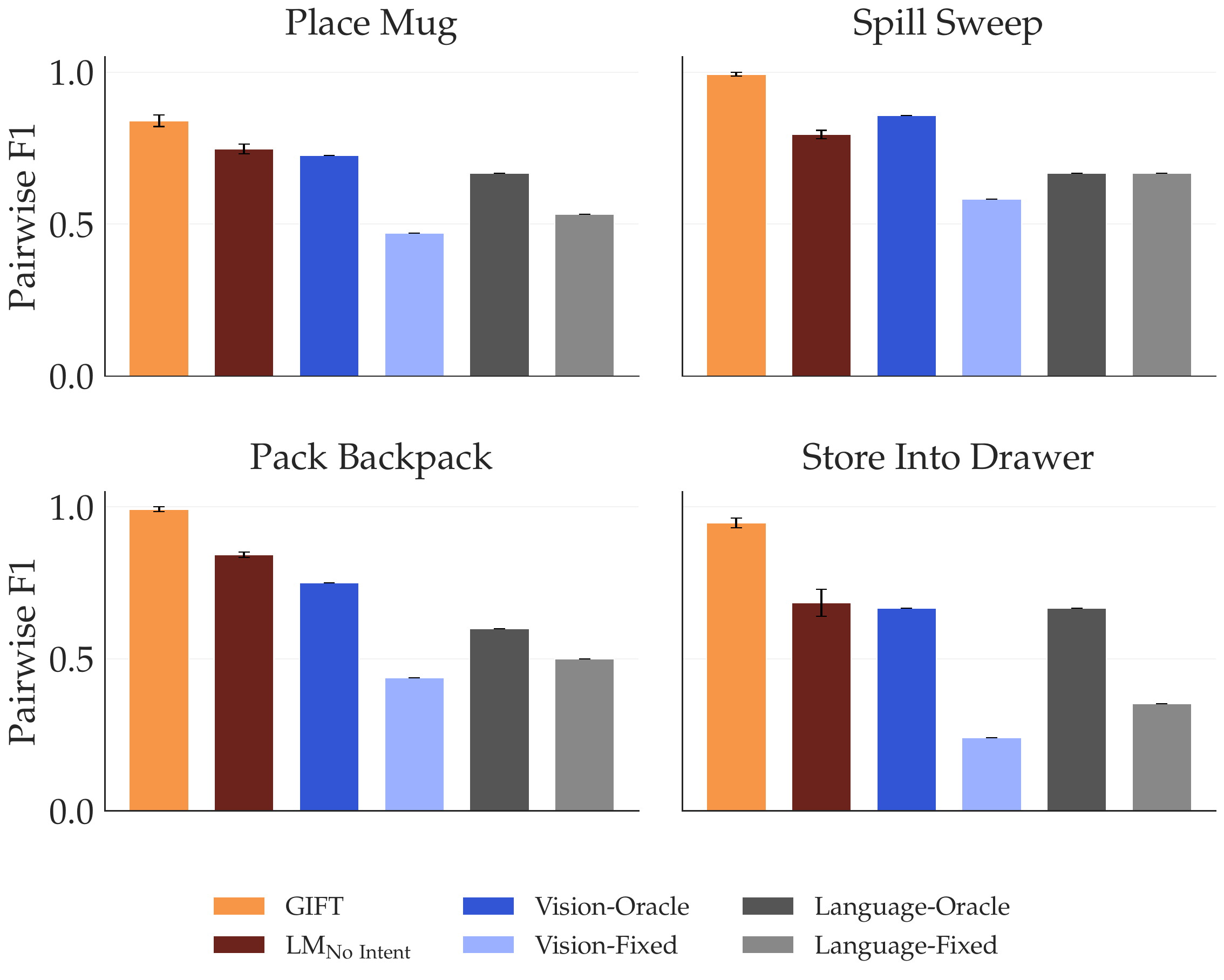}
    \vspace{-3mm}
    \caption{\textbf{Test-Time State Alignment F1 Score.} GIFT achieves a superior F1 score by aligning along intent, reducing confounds from superficial language/visual similarity.}
    \label{fig:f1}
    \vspace{-5mm}
\end{figure}

\begin{figure*}[t]
  \centering
  \begin{minipage}{0.49\textwidth}
    \centering
    \includegraphics[width=\linewidth]{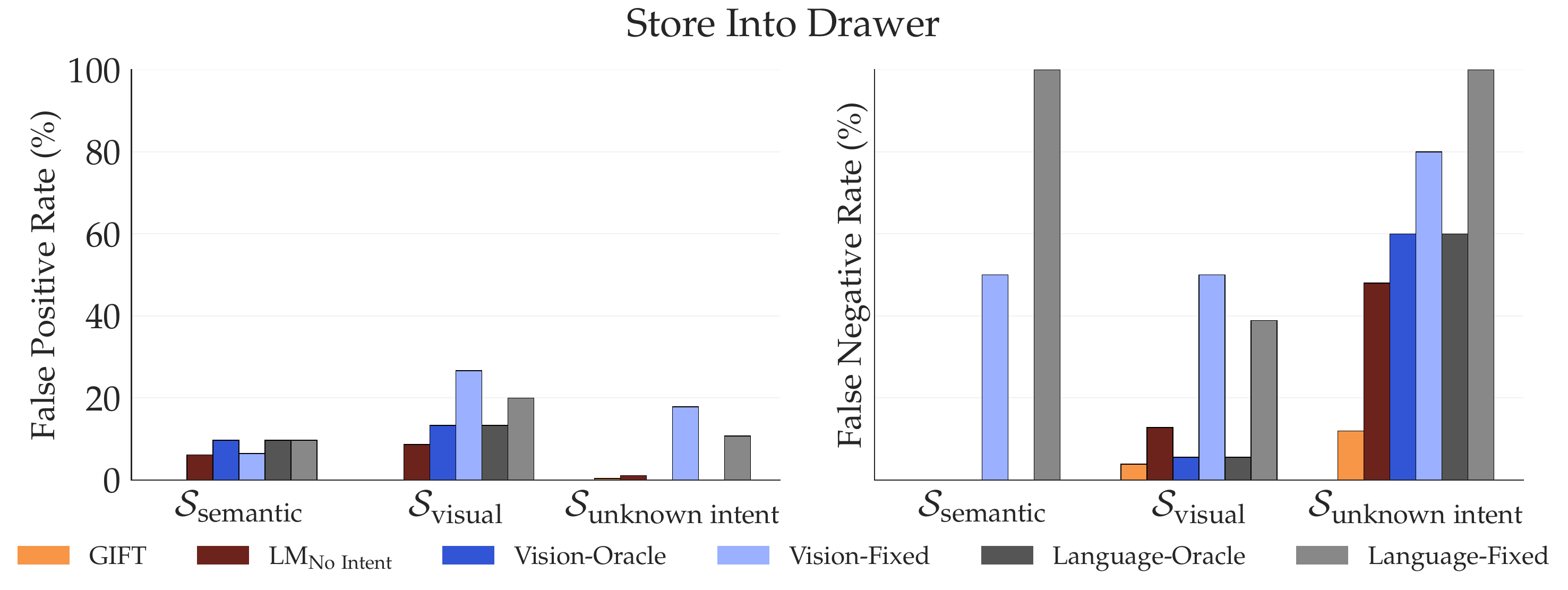}

  \end{minipage}\hfill
  \begin{minipage}{0.49\textwidth}
    \centering
    \includegraphics[width=\linewidth]{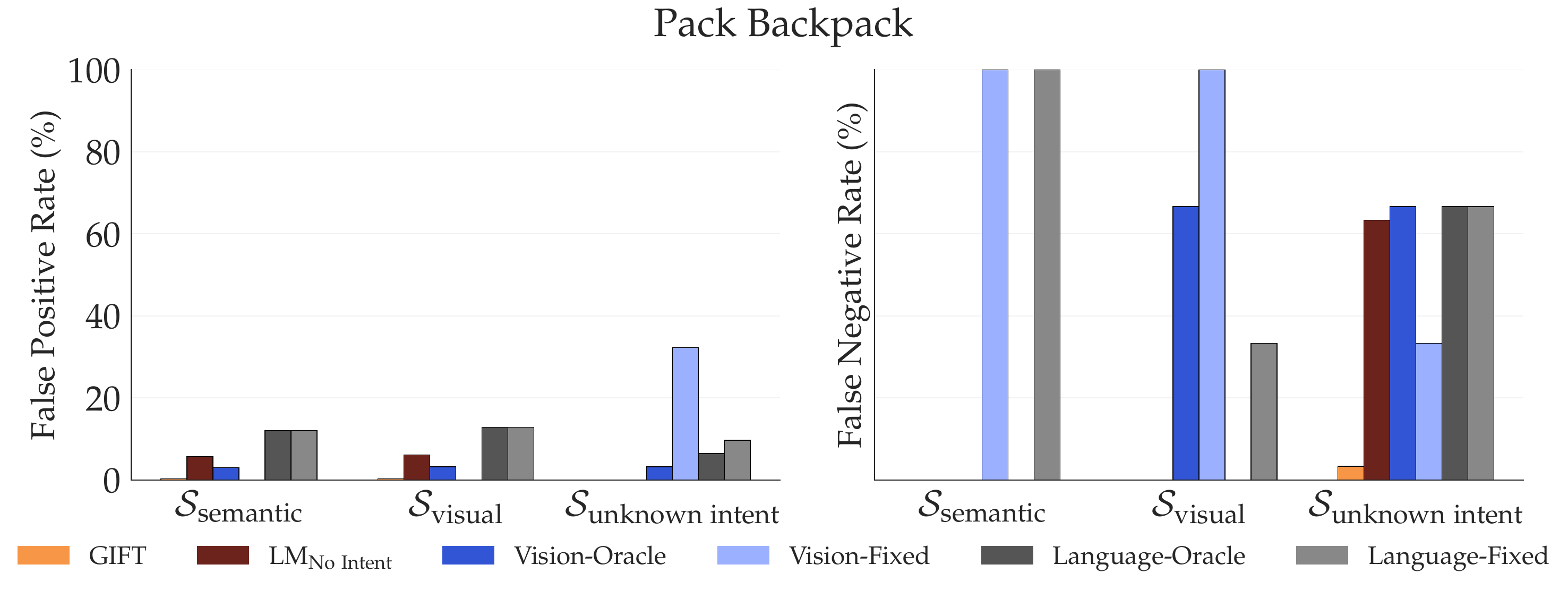}

  \end{minipage}
   \caption{\textbf{FP/FN (\%) on the Confounding States, $\mathcal{S}_{\mathrm{conf}}$.} GIFT remains low across categories by using intent-relevance. On the other hand, the oracle baselines merely retune thresholds and trade errors across confounds. Thresholding cannot correct a misaligned similarity signal, and therefore they show high errors. GIFT's ablation, LM$_{\mathrm{No\; Intent}}$, is less performant due to not recognizing which test states are intent relevant; leading to a high FN a rate.}
  \label{fig:patho_fpn_combined}
  \vspace{-5mm}
\end{figure*}

\subsection{Effectiveness Compared to Baselines}

\smallskip\noindent\textbf{Experimental Setup.}
\textbf{RQ1} aims to investigate if GIFT's intent-based similarity leads to better test-time reward inference compared to low-level visual or language similarity. We compare GIFT to three baseline similarity methods:
\begin{enumerate}
    \item \textbf{Vision}, $\mathcal{K}(\cdot,\cdot|\omega_{\text{vis}})$,
    by using the cosine similarity of DINO embeddings \cite{DINO} over images of scene objects. To mitigate the impact of noise from object appearances, we used Stable Diffusion~\cite{StableDiffusion} to generate three object images and compute their average DINO embedding.
    \item \textbf{Language}, $\mathcal{K}(\cdot,\cdot|\omega_{\text{lang}})$, by using the cosine similarity of BERT embeddings over textual descriptions of the objects in the scene.
    \item \textbf{LM$_{\mathrm{No\; Intent}}$}, $\mathcal{K}(\cdot,\cdot|\omega_{\text{LM}})$, parameterized by prompting an LM to directly map components of the test state to the training states as detailed in Sec.~\ref{sec:gift_implementation}, but without providing an intent learned from demonstrations. This baseline is an ablation of the intention conditioning mechanism of GIFT.  
\end{enumerate}

We use the same align-and-score procedure for each similarity method, where $\omega$ denotes the alignment conditioning signal (with GIFT using $\omega_{\text{GIFT}}=\hat{\omega}$). We compare $\omega \in \{\omega_{\text{vis}},\, \omega_{\text{lang}},\, \omega_{\text{LM}},\, \omega_{\text{GIFT}}\}$ by aligning a test trajectory with the training set using:
\begin{equation*}
\widetilde{\mathcal{R}}_\theta(\xi_{\text{test}}|\,\omega)
\;\triangleq\;
\mathcal{R}_\theta\!\big(f_{\omega}(\xi_{\text{test}})\big).
\end{equation*}

To mitigate the impact of noise from sampling LMs, we computed $\omega = J_{\mathrm{LM}}$, then performed the alignment procedure $f_{\omega}(s')$ for $s' \in \mathcal{S}^{\mathrm{test}}$ ten times. The final aligned state, $s'$, was set as the mode of these repeated runs.


We evaluated \textbf{RQ1} by generating a mixture of human-preferred and non-preferred test trajectories, randomizing across start locations, goal locations, object types, and object placements. We report average \textit{win rate} \cite{SPLASH,context_matters} for pairs of trajectories $(\xi_i,\xi_j)$, defined as the accuracy of predicting which trajectory the human prefers; a correct prediction counts as $+1$ for a given pair, and ties count as $+0.5$. For each scene, we average over 250 unique trajectories and three random seeds. We formed the following hypothesis for RQ1:

\smallskip
\noindent\textbf{H1.} Intent-conditioned similarity achieves a higher win rate than visual, language and LM$_{\mathrm{No\; Intent}}$ similarity methods.
\smallskip

\noindent\textbf{Results.}
Fig.~\ref{fig:winrate} shows pairwise win rate for each of the algorithms that were evaluated across the four tasks. Unseen trajectory pairs show increased win rates for GIFT across all tasks. The largest margins appear in more complex tasks. \textbf{Pack Backpack} and \textbf{Store Into Drawer} showed 20\% improvement over the next best baseline similarity methods. \textbf{Place Mug} and \textbf{Sweep Spill} achieved 7\% improvement over the next best baselines.
This result indicates that high-level intent similarity leads to better generalization in unseen test-time scenarios.




\subsection{Failure Modes for Low-level Features}

\noindent\textbf{Experimental Setup.} \textbf{RQ2} examines where low-level vision or language similarity methods fail to generalize at test time. To inspect this in detail we compare the ability of different similarity methods to correctly map a test-time object to a ground-truth train-time object.

Each method maps a test state $s'$ to a training state $s\in\mathcal{S}^{\mathrm{train}}$ or flags it as \textit{distractor} via a \textit{similarity threshold} for each of the four tasks. 
We evaluate the resulting state alignment using three metrics: binary F1, false positives (FP), and false negatives (FN) against a set of ground-truth intent labels. We compare two variants for vision-based and language-based similarities: \textit{oracle} and \textit{non-oracle}. The \textit{oracle variant} selected a threshold that maximize F1 on the test set--requiring access to ground-truth labels--and represents an upper bound. The \textit{non-oracle variant} instead used a fixed threshold selected by computing the average similarity of $\mathcal{S}^{\mathrm{train}}$. A FP represents that a traning object aligned with an irrelevant test object (e.g. \textit{toothbrush} $\rightarrow$ \textit{paintbrush}). A FN represents that a relevant test object is aligned to a \textit{distractor} (e.g. \textit{molding clay} $\rightarrow$ \textit{distractor}).

To facilitate analysis, we define the following datasets of test states $\mathcal{S}_{\mathrm{(\cdot)}}\subset \mathcal{S}^{\mathrm{test}}$. $\mathcal{S}_{\mathrm{R}}$ contains a mix of intent-relevant states and straightforward negatives. We additionally define \textit{confounding subsets} for tasks 3 and 4, to test the robustness of our approach to the following confounding factors:
\textbf{language confounds} $\mathcal{S}_{\mathrm{lang}}$ (i.e., items with similar names, such as \textit{toothbrush} and \textit{paintbrush}),
\textbf{visual confounds} $\mathcal{S}_{\mathrm{vis}}$ (i.e., items with similar appearances, such as \textit{broomstick} and \textit{paintbrush}),
and \textbf{unknown-intent confounds} $\mathcal{S}_{\mathrm{unk}}$ (i.e., sets of items that may be grouped in multiple ways depending on high-level intent, such as \textit{metal paintbrush}, \textit{wooden pencil}, and \textit{screwdriver}; two possible intent groupings are \textit{metal items} or \textit{art supplies}). Items may belong to multiple confounding subsets, and we define their union as $\mathcal{S}_{\mathrm{conf}} = \mathcal{S}_{\mathrm{lang}} \cup \mathcal{S}_{\mathrm{vis}} \cup \mathcal{S}_{\mathrm{unk}}$. We report classification error metrics for $\mathcal{S}_{\mathrm{R}} \cup \mathcal{S}_{\mathrm{conf}}$ and  $\mathcal{S}_{\mathrm{(\cdot)}}$ for tasks 3 and 4 and $\mathcal{S}_{\mathrm{R}}$ for the other two. This experiment was conducted in our simulated environment. Our hypothesis for RQ2 is:


\smallskip
\noindent\textbf{H2.} GIFT will produce fewer false positives and false negatives on the different types of confounds than language, visual, or LM$_{\mathrm{No\; Intent}}$ baselines.
\smallskip

\noindent\textbf{Results.}
Fig.~\ref{fig:f1} shows GIFT's superior alignment performance across tasks, which explains the uplift over baselines in Fig.~\ref{fig:winrate}. On tasks 3 and 4, GIFT lowers both FP and FN on the confounded subsets compared to visual/language baselines (Fig.~\ref{fig:patho_fpn_combined}). Overall, while the oracle baselines were sometimes robust to one form of confound, they were susceptible to other types of confounds, reducing their generalizability. To understand this behavior, we examined item similarity based on low-level visual and language-based features for the \textbf{Pack Backpack \textit{with Art Supplies}} ($\omega^*_{3a}$) intent and the \textbf{Store \textit{Valuables} Into Drawer} intent ($\omega^*_{4a}$).

\begin{figure*}[t]
    \centering
    \includegraphics[width=0.80\linewidth]{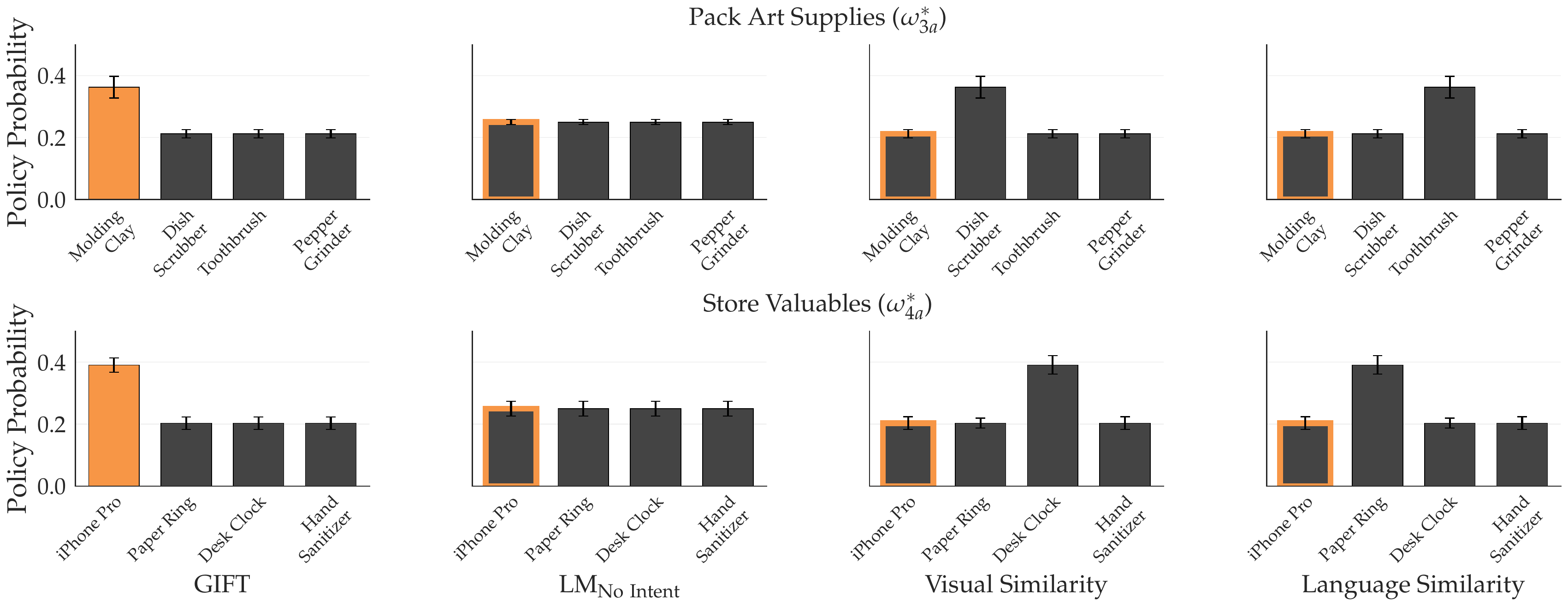}
    \vspace{-4mm}
    \caption{\textbf{Real World Behavior.} By utilizing intent, GIFT planned behavior that was more aligned with the human preferring to pack \textit{art supplies} and store \textit{valuables}. The x-axis labels correspond with test trajectories that have the Franka arm store/pack those unseen items. We randomly sampled sets of four trajectories corresponding to the four items and computed the resulting Boltzmann distribution for each induced reward along with the standard error.}
    \label{fig:real_world}
    \vspace{-4mm}
\end{figure*}

\begin{figure}[t]
    \centering
    \includegraphics[width=0.65\linewidth]{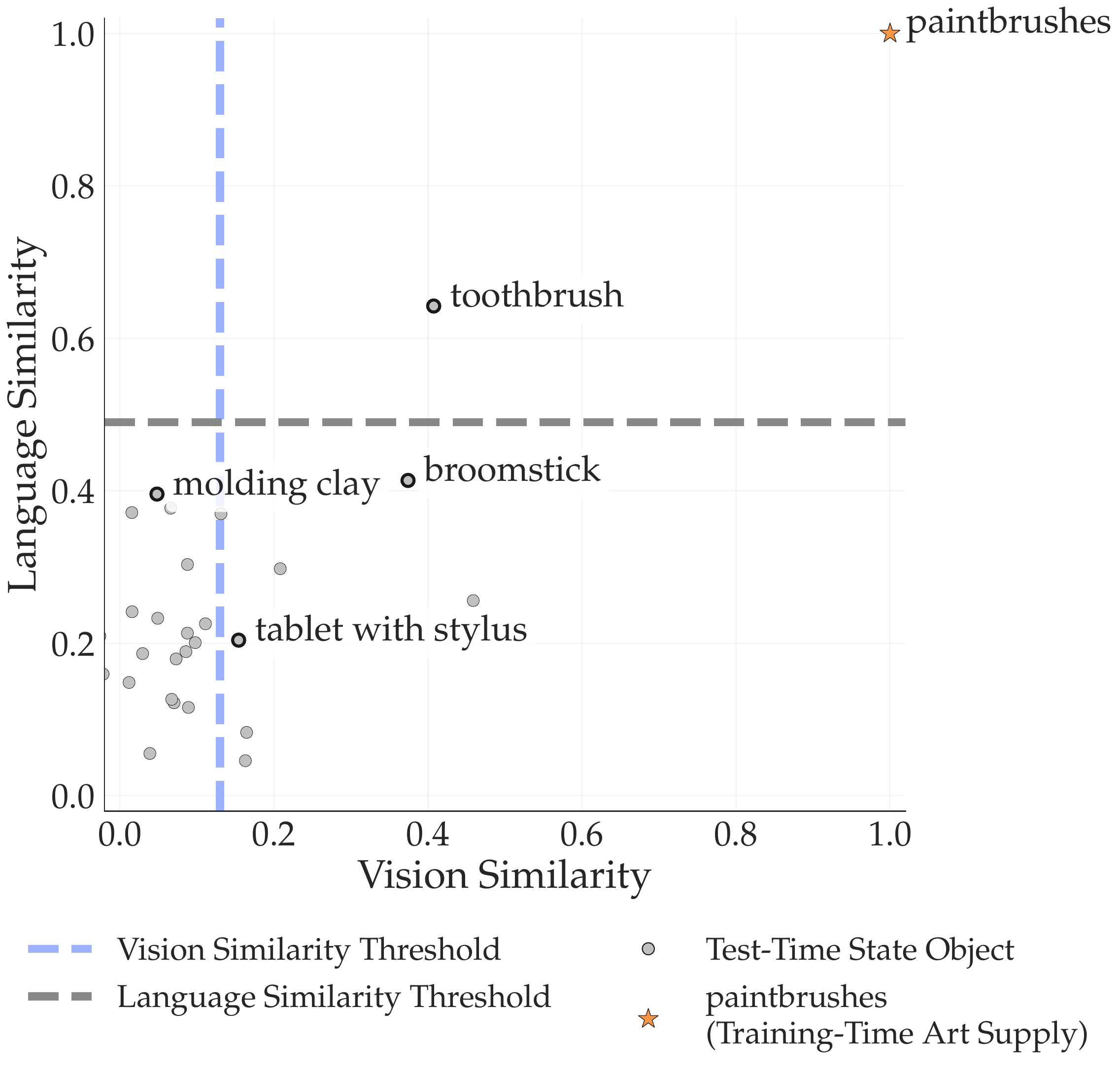}
    \vspace{-4mm}
    \caption{\textbf{Language--Vision Similarity Plot for Pack Backpack \textit{with Art Supplies}}. Similarities are measured relative to the training-time anchor (\textit{paintbrushes}) at $(1.0,1.0)$. Test objects such as the \textit{tablet with stylus} and \textit{molding clay} fall closer to \textit{distractor}, highlighting failure cases where relevant art supplies are confused with unrelated objects. Items like \textit{broomstick} and \textit{toothbrush} are incorrectly pulled toward \textit{paintbrushes}, showing how superficial visual or language similarity can lead to false positives. }
    \vspace{-5mm}
    \label{fig:2D_grid_art}
\end{figure}

\begin{figure}[t]
    \centering
    \includegraphics[width=0.65\linewidth]{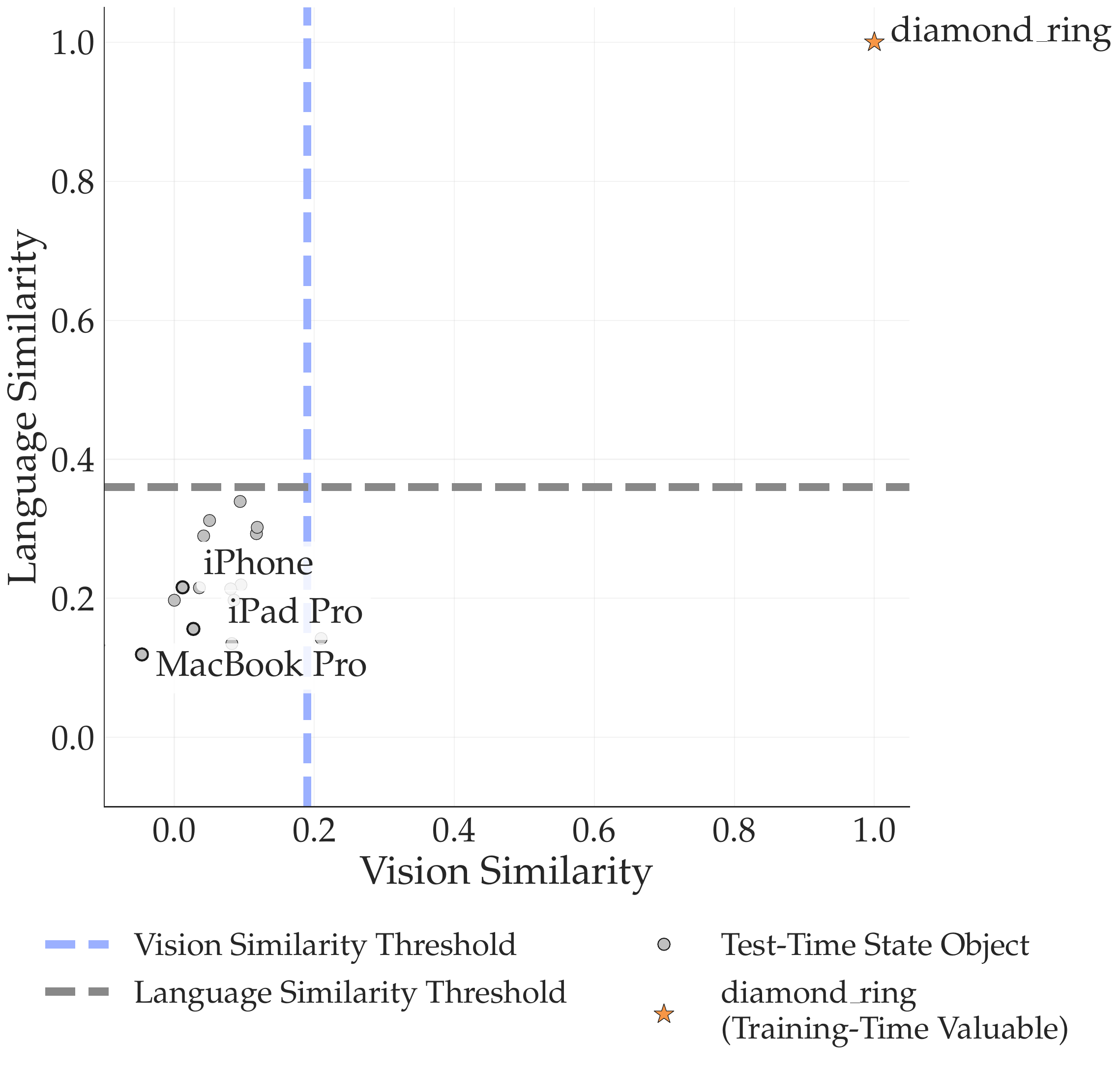}
    \vspace{-3mm}
    \caption{\textbf{Language--Vision Similarity Plot for \textit{Store Valuables}}. Similarities are measured relative to the training-time anchor (\textit{diamond ring}) at $(1.0, 1.0)$. Test objects such as the \textit{MacBook Pro} and \textit{iPad Pro} fall close to the anchor, reflecting correct generalization to unseen valuable items. In contrast, irrelevant objects like the \textit{paper ring} cluster further away, indicating proper separation between true valuables and distractors.}
    \label{fig:2D_grid_valuables}
    \vspace{-5mm}
\end{figure}

For $\omega^*_{3a}$ (Fig.~\ref{fig:2D_grid_art}), we observe that non art supplies like \textit{toothbrush} and \textit{broomstick} show high language and vision similarity with \textit{paintbrush}, yet items that are art supplies, like \textit{molding clay} and \textit{tablet with stylus}, show low similarity in these low-level visual and language feature spaces. Similarly, for $\omega^*_{4a}$ (Fig.~\ref{fig:2D_grid_valuables}), we observe that low-level visual and language similarity places the unvaluable \textit{paper ring} as highly similar to the valuable \textit{diamond ring}, whereas other valuable items such as \textit{MacBook Pro} and \textit{iPad Pro} are dissimilar to diamond ring. These results underscore the shortcomings of measuring similarity based on low-level features.


\subsection{Real World Experiments}

\noindent\textbf{Environmental Setup.} Our simulated experiments generated datasets of human-preferred trajectories given a ground truth reward function, but real-world humans may provide sub-optimal demonstrations. \textbf{RQ3} investigates whether LMs can infer intent from \textit{real-world} demonstrations from the Franka Emika Panda robot. We recreated two of the simulated tasks using Franka with a set of held-out physical objects. We then evaluated each method by repeatedly sampling a small candidate set of physically executable trajectories--each trajectory manipulating a different held-out object--and converting the resulting aligned-reward scores into a Boltzmann distribution over the candidates. For GIFT, we used intents inferred from Franka's trajectories, $J_{\mathrm{LM}}(\mathcal{D}^{\mathrm{Franka}}_{\mathrm{H}}, \mathcal{D}^{\mathrm{Franka}}_{\mathrm{\neg H}})$, to evaluate if the intents inferred from real-world demonstrations could facilitate robust reward learning. We developed the following hypothesis for RQ3:

\smallskip
\noindent\textbf{H3.} Intents inferred from real-world demonstrations will enable GIFT to plan behavior on physical robots that is better aligned with human preferences.
\smallskip

\noindent\textbf{Results.} In the Franka domain, we see similar success as in simulation. Planning with high-level intents produced behavior that is more aligned with human preferences on held-out objects than low-level visual features. In Fig.~\ref{fig:real_world}, GIFT correctly identified that the human prefers packing the \textit{molding clay} (an art supply) and storing the valuable \textit{iPhone Pro} and correctly treats the confounding held-out objects as distractors. As discussed in RQ2, the baselines succumb to spurious correlations in low-level features.   


\section{Conclusion}


GIFT reframes test-time reward reuse as an intent-conditioned alignment problem: instead of comparing states along low-level visual or language features, we align unseen states to behaviorally equivalent training states using an intent signal inferred from demonstrations. Across four tabletop tasks and more than 50 unseen objects, GIFT achieved consistent gains in pairwise win rate and lower FP/FN on confounded states versus DINO/BERT baselines and directly using an LM without inferring intent. These improvements were demonstrated in both a simulated JACO robot, aand on a physical 7-DoF Franka robot. In conclusion, by shifting comparisons from low-level visual or language features to higher-level representations of intent, GIFT enables robots to generalize reward functions to unseen test states.

\smallskip\noindent\textbf{Limitations and Future Work.} Limitations point to several next steps. First, performance depends on the quality of the inferred intent; LMs may unpredictably vary the level of abstraction of inferred intents leading to incorrect measures of similarity, and we do not yet calibrate confidence or abstain when uncertain. Future work includes learning an uncertainty-aware intent kernel, that can refine estimates of $\omega$ through additional user queries. Second, our experiments assumed symbolic descriptors for objects and scenes. In practice, these descriptors might be noisy and cause unexplored failure modes. Finally, because LMs can hallucinate and amplify training-set biases, deployment should include guardrails to mitigate potential harms.



\bibliographystyle{IEEEtran}
\bibliography{IEEEabrv, HRI}

@article{huang2022zeroshot,
  title   = {Language Models as Zero-Shot Planners: Extracting Actionable Knowledge for Embodied Agents},
  author  = {Huang, Wenlong and Abbeel, Pieter and Pathak, Deepak and Mordatch, Igor},
  journal = {arXiv preprint arXiv:2201.07207},
  year    = {2022}
}

@article{ma2023eureka,
  title   = {Eureka: Human-Level Reward Design via Coding Large Language Models},
  author  = {Ma, Yecheng Jason and Liang, William and Wang, Guanzhi and others},
  journal = {arXiv preprint arXiv:2310.12931},
  year    = {2023}
}

@inproceedings{peng2024algae,
  title     = {Adaptive Language-Guided Abstraction from Contrastive Explanations},
  author    = {Peng, Andi and Bobu, Andreea and Li, Belinda Z. and Sucholutsky, Ilia and Kumar, Nishanth and Shah, Julie and Andreas, Jacob},
  booktitle = {Conference on Robot Learning},
  year      = {2024}
}

@article{li2023lampp,
  title   = {LAMPP: Language Models as Probabilistic Priors for Perception and Action},
  author  = {Li, Belinda Z. and Chen, William and Sharma, Pratyusha and Andreas, Jacob},
  journal = {arXiv preprint arXiv:2302.02801},
  year    = {2023}
}

@inproceedings{FERL,
  title     = {Feature Expansive Reward Learning: Rethinking Human Input},
  author    = {Bobu, Andreea and Wiggert, Marius and Tomlin, Claire and Dragan, Anca D.},
  booktitle = {ACM/IEEE International Conference on Human-Robot Interaction},
  year      = {2021}
}

@article{GPT4,
  title={Gpt-4o system card},
  author={Hurst, Aaron and Lerer, Adam and Goucher, Adam P and Perelman, Adam and Ramesh, Aditya and Clark, Aidan and Ostrow, AJ and Welihinda, Akila and Hayes, Alan and Radford, Alec and others},
  journal={arXiv preprint arXiv:2410.21276},
  year={2024}
}

@article{Franka,
  author    = {Sami Haddadin},
  title     = {The Franka Emika Robot: A Standard Platform in Robotics Research [Survey]},
  journal   = {IEEE Robotics and Automation Magazine},
  volume    = {31},
  number    = {4},
  pages     = {136--148},
  year      = {2024},
  doi       = {10.1109/MRA.2024.3451788}
}

@article{LearningStructure,
  title   = {Inducing Structure in Reward Learning by Learning Features},
  author  = {Bobu, Andreea and Wiggert, Marius and Tomlin, Claire and Dragan, Anca D.},
  journal = {The International Journal of Robotics Research},
  volume  = {41},
  number  = {5},
  pages   = {497--518},
  year    = {2022},
  doi     = {10.1177/02783649221078031}
}

@article{ElicitingHumanPref,
  title={Eliciting human preferences with language models},
  author={Li, Belinda Z and Tamkin, Alex and Goodman, Noah and Andreas, Jacob},
  journal={arXiv preprint arXiv:2310.11589},
  year={2023}
}

@inproceedings{christiano2017prefs,
  title     = {Deep Reinforcement Learning from Human Preferences},
  author    = {Christiano, Paul and Leike, Jan and Brown, Tom and others},
  booktitle = {Advances in Neural Information Processing Systems},
  year      = {2017}
}

@article{sumers2021lingrewards,
  title   = {Learning Rewards from Linguistic Feedback},
  author  = {Sumers, Theodore R. and Ho, Mark K. and Hawkins, Robert D. and Narasimhan, Karthik and Griffiths, Thomas L.},
  journal = {Proceedings of the AAAI Conference on Artificial Intelligence},
  year    = {2021}
}

@article{mehta2024unified,
author = {Mehta, Shaunak A. and Losey, Dylan P.},
title = {Unified Learning from Demonstrations, Corrections, and Preferences during Physical Human–Robot Interaction},
year = {2024},
issue_date = {September 2024},
publisher = {Association for Computing Machinery},
address = {New York, NY, USA},
volume = {13},
number = {3},

doi = {10.1145/3623384},
journal = {J. Hum.-Robot Interact.},
month = aug,
articleno = {39},
numpages = {25}
}

@INPROCEEDINGS{CLEA,
  author={Dennler, Nathaniel and Nikolaidis, Stefanos and Matarić, Maja},
  booktitle={2025 20th ACM/IEEE International Conference on Human-Robot Interaction (HRI)}, 
  title={Contrastive Learning from Exploratory Actions: Leveraging Natural Interactions for Preference Elicitation}, 
  year={2025},
  volume={},
  number={},
  pages={778-788},
  doi={10.1109/HRI61500.2025.10974136}}

@inproceedings{myers2022learning,
  title={Learning multimodal rewards from rankings},
  author={Myers, Vivek and Biyik, Erdem and Anari, Nima and Sadigh, Dorsa},
  booktitle={Conference on robot learning},
  pages={342--352},
  year={2022},
  organization={PMLR}
}

@article{context_matters,
  title={Context Matters: Learning Generalizable Rewards via Calibrated Features},
  author={Forsey-Smerek, Alexandra and Shah, Julie and Bobu, Andreea},
  journal={arXiv preprint arXiv:2506.15012},
  year={2025}
}

@article{SPLASH,
  title={SPLASH! Sample-efficient Preference-based inverse reinforcement learning for Long-horizon Adversarial tasks from Suboptimal Hierarchical demonstrations},
  author={Crowley, Peter and Serlin, Zachary and Paine, Tyler and Mann, Makai and Benjamin, Michael and Belta, Calin},
  journal={arXiv preprint arXiv:2507.08707},
  year={2025}
}

@article{BERT,
  author    = {Jacob Devlin and Ming-Wei Chang and Kenton Lee and Kristina Toutanova},
  title     = {BERT: Pre-training of Deep Bidirectional Transformers for Language Understanding},
  journal   = {arXiv preprint arXiv:1810.04805},
  year      = {2018}
}

@inproceedings{DINO,
  author    = {Mathilde Caron and Hugo Touvron and Ishan Misra and Hervé Jégou and Julien Mairal and Piotr Bojanowski and Armand Joulin},
  title     = {Emerging Properties in Self-Supervised Vision Transformers},
  booktitle = {Proceedings of the IEEE/CVF International Conference on Computer Vision (ICCV)},
  year      = {2021}
}

@article{DVDLearningGeneralizableRewards,
  title={Learning generalizable robotic reward functions from" in-the-wild" human videos},
  author={Chen, Annie S and Nair, Suraj and Finn, Chelsea},
  journal={arXiv preprint arXiv:2103.16817},
  year={2021}
}

@article{semantically_controllabe_augmentations,
  title={Semantically controllable augmentations for generalizable robot learning},
  author={Chen, Zoey and Mandi, Zhao and Bharadhwaj, Homanga and Sharma, Mohit and Song, Shuran and Gupta, Abhishek and Kumar, Vikash},
  journal={The International Journal of Robotics Research},
  pages={02783649241273686},
  year={2024},
  publisher={SAGE Publications Sage UK: London, England}
}

@inproceedings{ROBO_ABC,
  title={Robo-abc: Affordance generalization beyond categories via semantic correspondence for robot manipulation},
  author={Ju, Yuanchen and Hu, Kaizhe and Zhang, Guowei and Zhang, Gu and Jiang, Mingrun and Xu, Huazhe},
  booktitle={European Conference on Computer Vision},
  pages={222--239},
  year={2024},
  organization={Springer}
}

@inproceedings{TargetProjection,
  title={Generalization on unseen domains via inference-time label-preserving target projections},
  author={Pandey, Prashant and Raman, Mrigank and Varambally, Sumanth and Ap, Prathosh},
  booktitle={Proceedings of the IEEE/CVF Conference on Computer Vision and Pattern Recognition},
  pages={12924--12933},
  year={2021}
}

@inproceedings{LearningToIdentifyNewObjects,
  title={Learning to identify new objects},
  author={Sun, Yuyin and Bo, Liefeng and Fox, Dieter},
  booktitle={2014 IEEE International Conference on Robotics and Automation (ICRA)},
  pages={3165--3172},
  year={2014},
  organization={IEEE}
}

@misc{StableDiffusion,
  title        = {High-Resolution Image Synthesis with Latent Diffusion Models},
  author       = {Robin Rombach and Andreas Blattmann and Dominik Lorenz and Patrick Esser and Björn Ommer},
  year         = {2021},
  eprint       = {2112.10752},
  archivePrefix= {arXiv},
  primaryClass = {cs.CV}
}

@book{puterman2014markov,
  author    = {Martin L. Puterman},
  title     = {Markov Decision Processes: Discrete Stochastic Dynamic Programming},
  series    = {Wiley Series in Probability and Statistics},
  publisher = {John Wiley \& Sons},
  address   = {New York},
  year      = {1994},
  isbn      = {978-0-471-61977-2},
  pages     = {1--649}
}

@inproceedings{abbeel2004apprentice,
    title={Apprenticeship {L}earning via {I}nverse {R}einforcement {L}earning},
    author={Abbeel, Pieter and Andrew Y. Ng},
    booktitle={Proceedings of the International Conference on Machine learning},
    year={2004}
}

@inproceedings{ziebart2008maximum,
  title={Maximum entropy inverse reinforcement learning.},
  author={Ziebart, Brian D and Maas, Andrew L and Bagnell, J Andrew and Dey, Anind K and others},
  booktitle={Aaai},
  volume={8},
  pages={1433--1438},
  year={2008},
  organization={Chicago, IL, USA}
}

@inproceedings{finn2016guided,
  title={Guided cost learning: Deep inverse optimal control via policy optimization},
  author={Finn, Chelsea and Levine, Sergey and Abbeel, Pieter},
  booktitle={International conference on machine learning},
  pages={49--58},
  year={2016},
  organization={PMLR}
}

@article{shimodaira2000improving,
  title={Improving predictive inference under covariate shift by weighting the log-likelihood function},
  author={Shimodaira, Hidetoshi},
  journal={Journal of statistical planning and inference},
  volume={90},
  number={2},
  pages={227--244},
  year={2000},
  publisher={Elsevier}
}

@inproceedings{rosid,
  title={The rosid tool: Empowering users to design multimodal signals for human-robot collaboration},
  author={Dennler, Nathaniel and Delgado, David and Zeng, Daniel and Nikolaidis, Stefanos and Matari{\'c}, Maja},
  booktitle={International Symposium on Experimental Robotics},
  pages={3--10},
  year={2023},
  organization={Springer}
}

@inproceedings{fu2018AIRL,
  title={Learning Robust Rewards with Adversarial Inverse Reinforcement Learning},
  author={Fu, Justin and Luo, Katie and Levine, Sergey},
  booktitle={International Conference on Learning Representations (ICLR)},
  year={2018}
}

@inproceedings{agrawal2022task,
  title={The Task Specification Problem},
  author={Agrawal, Pulkit},
  booktitle={Conference on Robot Learning},
  pages={1745--1751},
  year={2022},
  organization={PMLR}
}

@inproceedings{bobu2023SIRL,
  author       = {Andreea Bobu and
                  Yi Liu and
                  Rohin Shah and
                  Daniel S. Brown and
                  Anca D. Dragan},
  editor       = {Ginevra Castellano and
                  Laurel D. Riek and
                  Maya Cakmak and
                  Iolanda Leite},
  title        = {{SIRL:} Similarity-based Implicit Representation Learning},
  booktitle    = {Proceedings of the 2023 {ACM/IEEE} International Conference on Human-Robot
                  Interaction, {HRI} 2023, Stockholm, Sweden, March 13-16, 2023},
  pages        = {565--574},
  publisher    = {{ACM}},
  year         = {2023},
  doi          = {10.1145/3568162.3576989},
  bibsource    = {dblp computer science bibliography, https://dblp.org}
}

@article{amodei2016concrete,
  title={Concrete problems in AI safety},
  author={Amodei, Dario and Olah, Chris and Steinhardt, Jacob and Christiano, Paul and Schulman, John and Man{\'e}, Dan},
  journal={arXiv preprint arXiv:1606.06565},
  year={2016}
}

@inproceedings{bobu2024ARHR,
  author       = {Andreea Bobu and
                  Andi Peng and
                  Pulkit Agrawal and
                  Julie A. Shah and
                  Anca D. Dragan},
  editor       = {Dan Grollman and
                  Elizabeth Broadbent and
                  Wendy Ju and
                  Harold Soh and
                  Tom Williams},
  title        = {Aligning Human and Robot Representations},
  booktitle    = {Proceedings of the 2024 {ACM/IEEE} International Conference on Human-Robot
                  Interaction, {HRI} 2024, Boulder, CO, USA, March 11-15, 2024},
  pages        = {42--54},
  publisher    = {{ACM}},
  year         = {2024},
  doi          = {10.1145/3610977.3634987},
  bibsource    = {dblp computer science bibliography, https://dblp.org}
}

@inproceedings{bobu2018learning,
  title = 	 {Learning under Misspecified Objective Spaces},
  author = 	 {Bobu, A. and Bajcsy, A. and Fisac, J. F. and Dragan, A. D.},
  booktitle = {Conference on Robot Learning (CoRL)},
  year = 	 {2018}
}

@inproceedings{bajcsy2018learning,
  title={Learning from physical human corrections, one feature at a time},
  author={Bajcsy, Andrea and Losey, Dylan P and O'Malley, Marcia K and Dragan, Anca D},
  booktitle={Human Robot Interaction},
  year={2018}
}

@inproceedings{peng2023dfa,
  author       = {Andi Peng and
                  Aviv Netanyahu and
                  Mark K. Ho and
                  Tianmin Shu and
                  Andreea Bobu and
                  Julie Shah and
                  Pulkit Agrawal},
  title        = {Diagnosis, Feedback, Adaptation: {A} Human-in-the-Loop Framework for
                  Test-Time Policy Adaptation},
  booktitle    = {International Conference on Machine Learning, {ICML} 2023, 23-29 July
                  2023, Honolulu, Hawaii, {USA}},
  series       = {Proceedings of Machine Learning Research},
  volume       = {202},
  pages        = {27630--27641},
  publisher    = {{PMLR}},
  year         = {2023},
  bibsource    = {dblp computer science bibliography, https://dblp.org}
}

@inproceedings{peng2024plga,
author = {A. Peng and A. Bobu and B. Z. Li and T. Sumers and I. Sucholutsky and N. Kumar and T. Griffiths and J. Shah},
title = {Preference-Conditioned Language-Guided Abstraction},
year = {2024},
isbn = {979840079322},
doi = {10.1145/3610977.3634930},
booktitle = {Proceedings of the 2024 ACM/IEEE International Conference on Human-Robot Interaction},
series = {HRI '23}
}

@misc{forseysmerek2025contextmatterslearninggeneralizable,
      title={Context Matters: Learning Generalizable Rewards via Calibrated Features}, 
      author={Alexandra Forsey-Smerek and Julie Shah and Andreea Bobu},
      year={2025},
      eprint={2506.15012},
      archivePrefix={arXiv},
      primaryClass={cs.RO},
      
}

@MISC{coumans2019,
author =   {Erwin Coumans and Yunfei Bai},
title =    {PyBullet, a Python module for physics simulation for games, robotics and machine learning},
year = {2016--2019}
}

\onecolumn
\section*{Appendix}

\subsection{Alignment Error Bound}\label{sec:error_bound}
A natural concern is whether alignment could arbitrarily distort reward evaluation on novel test states. To address this, we show that under a mild smoothness assumption, the error introduced by alignment is bounded by the degree of dissimilarity \(1-\mathcal{K}\).

Suppose the per-step reward is $L$-Lipschitz with respect to the intent-conditioned distance $1-\mathcal{K}(\cdot,\cdot\mid\hat\omega)$: 
\begin{equation*} 
\begin{aligned} 
\big| \mathcal{R}_\theta(s) - \mathcal{R}_\theta(s') \big| &\;\le\; L\,\big(1-\mathcal{K}\!\big( s,\, s'\mid \hat\omega \big)\big) \\ \Rightarrow
\quad \big| \mathcal{R}_\theta(\xi) - \mathcal{R}_\theta(\xi') \big| &\;\le\; L \sum_{t} \big(1-\mathcal{K}\!\big( s_t,\, s'_t\mid \hat\omega \big)\big). 
\end{aligned} 
\end{equation*} 
Then for any test trajectory $\xi'$, the error from evaluating its aligned $f_{\hat{\omega}}(\xi')$ is bounded by the cumulative dissimilarity:
\begin{equation*} 
\big| \widetilde{\mathcal{R}}_\theta(\xi'\mid\hat\omega) - \mathcal{R}_\theta(\xi') \big| \;\le\; L \sum_t \big(1-\mathcal{K}(f_{\hat{\omega}}(s'_t), s'_t \mid \hat{\omega})\big).
\end{equation*}

This result clarifies when a fixed reward remains meaningful under GIFT: as long as aligned states remain highly similar under the intent-conditioned kernel, the error is small.

The Lipschitz assumption can be satisfied by instantiating $\mathcal{K}$ via continuous similarity measures, such as embedding-based kernels or LLM confidence scores. While our implementation of the kernel is binary, the bound still gives some intuition: states aligned with high confidence ($\mathcal{K} \approx 1$) incur minimal error, while forced alignments to dissimilar states may produce unreliable rewards.

\subsection{Prompt Templates}\label{sec:prompt_templates}

In this section, we report prompts for the LMs used in our experiments. In our implementation, we had three significant LM calls. \textbf{Call 1:} The first LM call corresponds with $\hat{\omega}_{}=J_{\mathrm{LM}}(\mathcal{D}_{\mathrm{H}}, \mathcal{D}_{\mathrm{\neg H}})$. We implemented this by prompting the LM to determine the intent of the user, given the textualized demonstrations. \textbf{Call 2:} The second call was made to determine the axis of similarity that is relevant to the intent. Functionally, this call embodied $\mathcal{K}(\cdot,\cdot|\omega_{\text{GIFT}})$, with GIFT using $\omega_{\text{GIFT}}=\hat{\omega}$. \textbf{Call 3:} The third call was made to perform the intent-conditioned alignment: $f_{\hat\omega}(s') \;\triangleq\; \operatorname*{argmax}_{s\in \mathcal{S}^{\mathrm{train}}}\; \mathcal{K}\!\big(s,\,s'\mid \hat\omega\big),$ where $s'$ is a novel state. The operator maps $s'$ to the nearest intent-equivalent training state identified by the kernel $\mathcal{K}$. Note that we implemented LM$_{\mathrm{No\; Intent}}$ by forgoing Calls 1 and 2 and using Call 3 with $\omega_{\text{LM}}$ (common sense LM reasoning without intent). \\

\textbf{Call 1:}\\

\begin{promptbox}{System Prompt: Infer intent \ensuremath{\hat{\omega}=J_{\mathrm{LM}}(\mathcal{D}_{\mathrm{H}}, \mathcal{D}_{\neg \mathrm{H}})}}
You are a language model tasked with identifying human-like motivations from robot demonstrations. The demonstrations include default and human-preferred trajectories performed by a robotic arm. Analyze the trajectories to determine the higher-level task goal that the human is trying to accomplish, along with the relevant movement features. Consider not only how the movement is executed but also what the objects being moved represent and their likely intended use. The motivation should reflect why the human is performing the task rather than just how the movement is executed (e.g., do not write something like 'Minimize movement variability'). Ensure that your motivations are task-specific and ordered by decreasing likelihood. KEEP IN MIND THAT WE WANT SPECIFIC motivations/PREFERENCES. Do not answer with something generic such as 'The human prefers to avoid objects.' WHAT KIND OF OBJECTS? WHAT IS THE MOTIVATION OR GOAL? BE AS SPECIFIC AS POSSIBLE. Ensure that the motivations are clearly named, and each relevant feature corresponds to a feature provided in the trajectory data. Your motivations should reflect underlying human intent and be framed in natural language.

You must produce a single Python dictionary with:
- 3 motivations, ordered by likelihood.
- Each motivation written as a full-sentence human preference (max 2 sentences).
- Each motivation mapped to a list of relevant feature names from the trajectory data.
- You may reuse features across multiple motivations, but you must not create new features.

 - Avoid referencing the objects and instead focus on their categories. For example, don't write 'the human prefers to avoid scalpels near peaches' instead write something like 'the human prefers to avoid sharp objects near fruit'The 'proximity' feature is maximized (set to 1.0) when the object pairs are close to each other, and minimized (0.0) when the objects are far apart or one of them is missing from the scene. In other words, proximity is the opposite of distance.The response must be a Python dictionary as a string, and nothing else.

Example Input:
Default trajectory:
Timestep 0: banana_proximity_to_scalpel: 0.11
Timestep 1: banana_proximity_to_scalpel: 0.42
...
Timestep 60: banana_proximity_to_scalpel: 0.85

Human-preferred trajectory:
Timestep 0: banana_proximity_to_scalpel: 0.45
...
Timestep 60: banana_proximity_to_scalpel: 0.14

Additional features:
- peach_proximity_to_suture
- banana_proximity_to_backpack

Example Output:
{
  "The human prefers to keep edible objects at a safe distance from sharp tools.": [
    "banana_proximity_to_scalpel",
    "peach_proximity_to_suture",
  ],
  "The human prefers to organize items of the same category together for transport.": [
    "banana_proximity_to_scalpel",
    "peach_proximity_to_suture",
  ],
  "The human avoids placing sharp tools near objects that may be handled directly.": [
    "banana_proximity_to_scalpel",
    "peach_proximity_to_suture",
  ]
}

Bad Output Example (What NOT to do):
{
  "The human prefers to minimize distance for efficiency.": ["objectA_to_objectB_proximity"],
  "The human wants things to be packed efficiently.": ["objectC_to_objectD_proximity"],
  "The human reduces object-to-bag proximity for optimized storage.": ["objectE_to_objectF_proximity"]
}
These responses are too generic, paraphrase the same vague idea, and fail to explain what type of objects are involved or why the behavior reflects a specific human intent. Do not simply reword 'minimize distance' -- you must infer the kind of objects and what goal they help accomplish.
\end{promptbox}

\begin{promptboxOrange}{User Prompt: Infer intent \ensuremath{\hat{\omega}=J_{\mathrm{LM}}(\mathcal{D}_{\mathrm{H}}, \mathcal{D}_{\neg \mathrm{H}})}}
You are a language model tasked with understanding robotic behaviors to infer human-like motivations and identify relevant features associated with each motivation. <TASK_STRING> The goal is to generalize these behaviors for unseen tasks. The provided data includes default trajectories and human-preferred trajectories.

Approach this task by considering how humans might act in similar scenarios and what factors influence their choices. We aim to uncover the underlying reasons behind robotic actions and human-like preferences purely from movement patterns. In this task, the Jaco robotic arm moves objects in an environment following different movement patterns. You will analyze variations in these movement sequences to determine possible patterns and associations. Answer the following questions:
1) What could be the broader goal of the robotic arm's actions?
2) How do variations in the arm's movement provide clues about preferences or motivations?
3) Which features highlight key differences between the human-prefered and default trajectories?
4) How do the properties and functions of the objects influence the trajectory??
5) Based on the motivations of the human, what would be a more general definition for the features?
6) What do the objects across the human-prefered demonstrations have in common? What do they tell you about the human's motivation?

Analyze the trajectories to determine the higher-level task goal that the human is trying to accomplish, along with the relevant movement features. Consider not only how the movement is executed but also what the objects being moved represent and their likely intended use. The motivation should reflect why the human is performing the task rather than just how the movement is executed (e.g., do not write something like 'Minimize movement variability'). Ensure that your motivations are task-specific and ordered by decreasing likelihood. KEEP IN MIND THAT WE WANT SPECIFIC motivations/PREFERENCES. Do not answer with something generic such as 'The human prefers to avoid objects.' WHAT KIND OF OBJECTS? WHAT IS THE MOTIVATION OR GOAL? BE AS SPECIFIC AS POSSIBLE. Ensure that the motivations are clearly named, and each relevant feature corresponds to a feature provided in the trajectory data. Furthermore, the motivations should be ordered by decreasing liklihood. In other words, your first motivation should be the most probable. Give three possible motivations. Avoid referencing the objects and instead focus on their categories. For example, don't write 'the human prefers to avoid scalpels near peaches' instead write something like 'the human prefers to avoid sharp objects near fruit'.Your motivation or preference entry can be up to two sentences. YOU SHOULD NOT CREATE NEW FEATURES, but you may reuse features across motivations.

**In addition to listing the motivations and features, you must also identify relevant semantic similarities between the feature sets.** Each semantic similarity should describe why a group of features is similar in terms of the inferred motivation. Think in terms of conceptual properties rather than physical similarities. Try to write the semantic similarity as a question.

The output should be structured as two Python dictionary-like objects in the following format:

    motivations = {
        'motivation or preference 1': ['relevant_feature_1', 'relevant_feature_2', ...],
        'motivation or preference 2': ['relevant_feature_3', 'relevant_feature_4', ...],
        ...
    }

    semantic_similarities = {
        'semantic concept 1': ['relevant_feature_1', 'relevant_feature_2', ...],
        'semantic concept 2': ['relevant_feature_3', 'relevant_feature_4', ...],
        ...
    }

The following are the trajectories:

Default trajectory:
<DEFAULT_TRAJ_BLOCK(v)>     
Human-preferred trajectory:
<PREFERRED_TRAJ_BLOCK(v)>

Think carefully about the motivations and their corresponding features before generating the dictionary. Your motivations should not only focus on movement constraints but also consider the role of the objects involved. If an object is consistently moved towards a specific location, infer the purpose of that action rather than just how the movement was performed.
EXTREMELY IMPORTANT. WE WILL SAVE YOUR OUTPUTS AS A PYTHON STRING VARIABLE. WRITE YOUR RESPONSE AS A PYTHON VARIABLE. THIS WILL BE FED INTO ANOTHER LANGUAGE MODEL. DO NOT USE ANY LATEX OR BOLD FONT Before finalizing your response, ask yourself: 'Am I describing the task goal or just the way it was executed?' If the answer is only about movement, reframe it to focus on what the human wanted to achieve with the task. YOUR RESPONSE SHOULD CONTAIN THE DICTIONARY OBJECT(s) AT THE VERY END
\end{promptboxOrange} \\
\textbf{Call 2:}\\

\begin{promptbox}{System Prompt: Infer similarity axis \ensuremath{\mathcal{K}(\cdot,\cdot\mid \omega_{\text{GIFT}}),\ \omega_{\text{GIFT}}=\hat{\omega}}}
You are a language model tasked with identifying the conceptual grouping axis that unifies a set of objects. These objects come from robot demonstrations, where each feature is a distance between two named objects (e.g., 'banana_distance_to_scalpel'). The goal is to reason about *why* certain objects appear together given the motivation behind a human's actions. Using this reasoning, determine the **semantic axis**--a short, natural language question or label--that explains what unites them conceptually.

You will be given a dictionary mapping inferred human motivations to lists of relevant feature names. Group those object names based on **shared conceptual properties** (e.g., same category, same function, same usage constraint). For each group, label it with a **semantic axis**, written as a short yes/no question (e.g., 'is this a fruit?').

Your output must be a single Python dictionary, where:
- Each key is a semantic axis (a question string).
- Each value is a list of object names that belong to that axis.
- Do not invent object names that don't appear in the features.
- Do not refer to specific demonstrations, environments, or file structures.

You must NOT output any text except the final dictionary. No explanations, no headings, no comments.

Be precise and avoid vague or overly general labels like 'is this an object?'.
\end{promptbox}

\begin{promptboxOrange}{System Prompt: Infer similarity axis \ensuremath{\mathcal{K}(\cdot,\cdot\mid \omega_{\text{GIFT}}),\ \omega_{\text{GIFT}}=\hat{\omega}}}
You are given a dictionary of inferred human motivations and their relevant features:

<MOTIVATION_DICT_JSON>

Each feature in this dictionary is a distance relationship between two objects, such as 'objectA_distance_to_objectB'.

Your task is to extract all the **individual objects** that appear in the feature names and group them based on shared conceptual properties.
Each group should be labeled with a semantic axis--a short question or phrase that identifies what unifies those objects conceptually.
Do not repeat motivation names, do not describe the full feature names, and do not refer to specific environments.
Each object should appear in only one axis group.

Your output should be a single Python dictionary mapping each semantic axis to a list of object names.

Do not output anything except the final dictionary.
\end{promptboxOrange}\\

\textbf{Call 3:}\\

\begin{promptbox}{System Prompt: Intent-conditioned alignment \ensuremath{f_{\hat{\omega}}(s') \triangleq \operatorname*{argmax}_{s\in \mathcal{S}^{\mathrm{train}}}\ \mathcal{K}(s,s'\mid \hat{\omega})}}
You are an assistant that helps generalize object-level behaviors from a set of seen objects to a new set of unseen objects. In each interaction, you are given:
- A list of objects previously seen in the training environment.
- A new list of objects present in the test environment.
- Sometimes, additional context in the form of inferred human motivations and conceptual similarity axes.

Your job is to map each object in the unseen state to either:
1. A specific object in the seen state that it meaningfully corresponds to based on human intent and conceptual relevance.
2. The special token 'distractor' if the object is not relevant to the human's motivations or does not align with any known object roles.

You will always output a single Python dictionary, and nothing else. The dictionary must map **every** unseen object to either a seen object or 'distractor'. You are allowed to map numerous unseen elements to the same seen elements.

You may encounter two types of input conditions:
- **Full Information**: You are given motivations and semantic similarity groupings.

- **Blind**: No prior motivations or similarity axes are given.

### REQUIRED FORMAT:
- Do not include markdown or code blocks.
- Do not add commentary or explanation.
- Only output a Python dictionary as plain text.

If you cannot find a strong match for an unseen object, assign it as 'distractor'. BE SURE TO USE THE SEMANTIC SIMILARITIES (if they are given to you). This will help you generalize behaviors beyond the class of objects which the human demonstrations contain. The idea is to generalize behaviors which means surface-level similarities are not always relevant.
\end{promptbox}

\begin{promptboxOrange}{User Prompt: Intent-conditioned alignment \ensuremath{f_{\hat{\omega}}(s') \triangleq \operatorname*{argmax}_{s\in \mathcal{S}^{\mathrm{train}}}\ \mathcal{K}(s,s'\mid \hat{\omega})}}
[If we are testing GIFT:]
EXTREMELY IMPORTANT, USE THE INFERRED MOTIVATIONS AND SEMANTIC SIMILARITIES WHEN MAKING YOUR DECISIONS. There are two dictionaries, the first gives you the motivations and their relevant features that were inferred from the trajectories. The second dictionary tells you the relevant semantic similarities with respect to the motivations:
<BACKGROUND_CALL>

[Else (We are testing LM No Intent):]
VERY IMPORTANT: Ignore future references to the mention of inferred semantic similarities or motivations in the rest of this message. Use your common sense reasoning.

The previously seen state contains the following objects:
<SEEN_OBJECT_LIST>

The new unseen state contains the following objects:
<UNSEEN_OBJECT_LIST>

Your task is to generalize the features and roles of objects in the unseen state based on the inferred motivations and their relevant features provided above. Additionally, use the provided semantic similarities to ensure that mappings prioritize conceptually relevant properties rather than just surface-level object categories. This will allow the unseen demonstrations to be analyzed using the existing reward model. Specifically:
1. Propose replacements or generalizations for objects that have no direct equivalent, considering the motivations, their features, and their relevant similarities.
2. Identify objects in the unseen state that directly correspond to objects in the following seen state.
<SEEN_OBJECT_LIST>

Explain how the new unseen state could be altered directly to make it compatible as feature arguments for the seen domain/reward model. Your explanation should include specific changes to the state structure or feature definitions, taking into account the relevant semantic similarities.

Finally, produce a Python dictionary that maps objects in the unseen state to objects in the seen state. This dictionary should match the format used by our `spoof_unseen_trajectories` function. For example:

{
  "objA": "objX",
  "objB": "distractor",
  "objC": "objZ"
  ...
}

Mappings should be justified using the inferred semantic similarities and motivations (if given), ensuring that objects are matched based on relevant conceptual properties rather than just visual or categorical resemblance. If an unseen object directly corresponds to a known one in the seen domain, map it accordingly. Otherwise, mark it as 'distractor'.

EXTREMELY IMPORTANT. WE WILL SAVE YOUR OUTPUTS AS A PYTHON STRING VARIABLE. WRITE YOUR RESPONSE AS A PYTHON VARIABLE. THIS WILL BE FED INTO ANOTHER LANGUAGE MODEL. DO NOT USE ANY LATEX OR BOLD FONT
\end{promptboxOrange}

\end{document}